\documentclass[lettersize,journal]{IEEEtran}
\usepackage{amsmath,amsfonts}
\usepackage{algorithmic}
\usepackage{algorithm}
\usepackage{array}
\usepackage[caption=false,font=normalsize,labelfont=sf,textfont=sf]{subfig}
\usepackage{textcomp}
\usepackage{stfloats}
\usepackage{url}
\usepackage{verbatim}
\usepackage{graphicx}
\usepackage{cite}
\usepackage{duckuments}
\usepackage{orcidlink}
\usepackage[export]{adjustbox}
\usepackage{lipsum} 
\usepackage{multirow}
\usepackage{booktabs}
\usepackage{siunitx}

\usepackage{todonotes}
\usepackage{afterpage}

\newcommand{\pValuePL}{\mbox{$p\!<\!.001$}}

\newcommand{\anovaPL}[5]{\mbox{$(\mathrm{#1}~F(#2,#3)\!=\!#4,~p\!<\!.001)$}}

\newcommand{\robustANOVAName}{\!\!}

\begin{document}

\title{Situation Awareness for Driver-Centric Driving Style Adaptation}

\author{Johann~Haselberger\,\orcidlink{0000-0002-2458-3461}$^*$,
Bonifaz~Stuhr\,\orcidlink{0000-0001-5452-8618}$^*$,
Bernhard~Schick\,\orcidlink{0000-0001-5567-3913},
and~Steffen~Müller
% <-this % stops a space
\thanks{
  Manuscript received March 17, 2024. 
  % ; revised August 16, 2021. 
\textit{($^*$contributed equally to this work.)} \textit{(Corresponding author: Johann Haselberger.)}}
% <-this % stops a space
\thanks{Johann Haselberger and Steffen Müller are with the Technische Universität Berlin, 10623 Berlin, Germany (e-mail: johann.haselberger@hs-kempten.de; steffen.mueller@tu-berlin.de).

Bonifaz Stuhr, Bernhard Schick, and Johann Haselberger are with the the University of Applied Science Kempten, 87435 Kempten, Germany (e-mail: bonifaz.stuhr@hs-kempten.de; bernhard.schick@hs-kempten.de).
}}

% The paper headers
\markboth{~}%
{Shell \MakeLowercase{\textit{et al.}}: A Sample Article Using IEEEtran.cls for IEEE Journals}

\IEEEpubid{This work has been submitted to the IEEE for possible publication. Copyright may be transferred without notice, after which this version may no longer be accessible.}
% Remember, if you use this you must call \IEEEpubidadjcol in the second
% column for its text to clear the IEEEpubid mark.

\newcommand{\clineSpacing}{\noalign{\vspace{0.05cm}}}

\maketitle

\begin{abstract}
There is evidence that the driving style of an autonomous vehicle is important to increase the acceptance and trust of the passengers. 
The driving situation has been found to have a significant influence on human driving behavior.
However, current driving style models only partially incorporate driving environment information, limiting the alignment between an agent and the given situation.
Therefore, we propose a situation-aware driving style model based on different visual feature encoders pretrained on fleet data, as well as driving behavior predictors, which are adapted to the driving style of a specific driver.
Our experiments show that the proposed method outperforms static driving styles significantly and forms plausible situation clusters.
% , leading to high interpretability of the predictions.
Furthermore, we found that feature encoders pretrained on our dataset lead to more precise driving behavior modeling. 
In contrast, feature encoders pretrained supervised and unsupervised on different data sources lead to more specific situation clusters, which can be utilized to constrain and control the driving style adaptation for specific situations.
Moreover, in a real-world setting, where driving style adaptation is happening iteratively, we found the MLP-based behavior predictors achieve good performance initially but suffer from catastrophic forgetting.
In contrast, behavior predictors based on situation-dependent statistics can learn iteratively from continuous data streams by design.
Overall, our experiments show that important information for driving behavior prediction is contained within the visual feature encoder.
The dataset is publicly available at \href{https://huggingface.co/datasets/jHaselberger/SADC-Situation-Awareness-for-Driver-Centric-Driving-Style-Adaptation}{\textbf{huggingface.co/datasets/jHaselberger/SADC-Situation-Awareness-for-Driver-Centric-Driving-Style-Adaptation}}.
\end{abstract}


\section{Introduction}

\IEEEPARstart{A}{s} autonomous vehicle development advances, attention is shifting from technical realizability to achieving driving characteristics that are both comfortable and acceptable \cite{bellem2018comfort}.
A crucial aspect of perceived driving comfort is influenced by the driving style, playing a vital role in fostering trust and acceptance of autonomous vehicles \cite{ekman2019exploring,strauch2019real,carsten2019can,ramm2014first}.
Considerable evidence shows that a driving style adaptation towards the human driver could improve the acceptance of autonomous driving functions and mitigate uncertainties associated with their usage \cite{martinez2017driving, sun2017research, bruck2021investigation, drewitz2020towards,chen2019driving,pion2012fingerprint,van2018relation,sun2020intention,gkartzonikas2019have,buyukyildiz2017identification,inagaki2003adaptive,bar2011probabilistic,chu2017curve,karlsson2021encoding,phinnemore2021happy,ma2021drivers}.
The term "driving style" lacks a comprehensive and standardized definition \cite{itkonen2020characterisation,chu2020self,chen2021driving}; however, definitions commonly agree that driving style encompasses a collection of driving habits developed and refined by a driver \cite{elander1993behavioral, lajunen2011self,sagberg2015review,kleisen2011relationship,tement2022assessment}.
It is argued that drivers prefer a style similar to their own \cite{hasenjager2019survey,festner2016einfluss,griesche2016should,bolduc2019multimodel,sun2020exploring,hartwich2015drive,rossner2022also,dettmann2021comfort}.
\begin{figure}[t]
    \centering
    \small
    \begin{tabular}{cc}
        \includegraphics[width=.4\linewidth,valign=m]{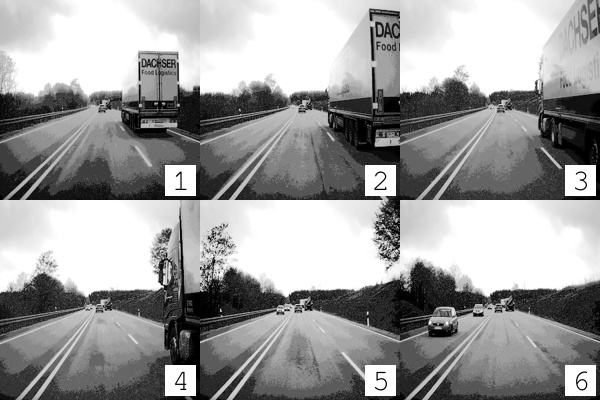} & 
        \includegraphics[width=.4\linewidth,valign=m]{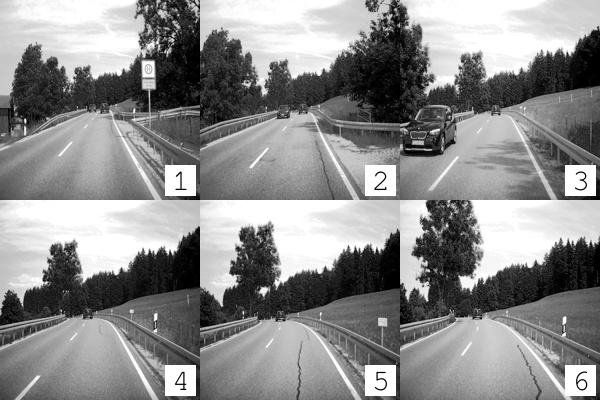}
         \\
        ~ &  ~ \\
        % \frame{
            \includegraphics[page=1,trim={1.2cm 0.65cm 1.22cm 0cm}, clip,width=.45\linewidth,valign=m]{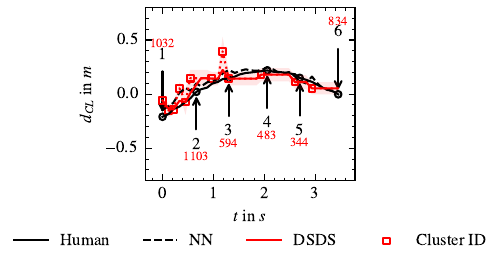}
        % }
         & 
        \includegraphics[page=1,trim={1.2cm 0.65cm 1.22cm 0cm}, clip,width=.45\linewidth,valign=m]{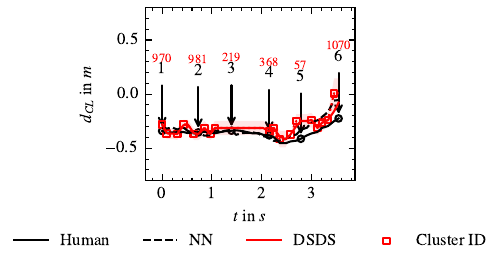} \\
        % I &  I \\
        ~ & ~ \\[-0.2cm]
        \multicolumn{2}{c}{
            % \frame{
                \includegraphics[page=1,trim={0cm 0.15cm 0.2cm 3.8cm}, clip,width=.7\linewidth,valign=m]{assets/sit_plots/001_824.pdf}
                % }
        } 
    \end{tabular}
    \caption{
        Distance to lane center predictions of our proposed neural-network-based driving style model (NN) and the driving situation clustering approach (DSC) for two specific scenarios.
        The top row shows images of the driving situation in chronological order, and the bottom row shows the predicted trajectories and the recorded human behavior.
        Red squares denote a change in the situation cluster identified by the DSC approach.
        Corresponding images and their respective cluster IDs are annotated with arrows.
        }
\label{fig:introFigure}
\end{figure}
Current driver models or driving functions, however, depict an average driver with static parameters \cite{hasenjager2019survey, chu2020self,gao2020personalized}, lacking adaptation to individual drivers \cite{ponomarev2019adaptation,rosenfeld2012learning,rosenfeld2012towards,karlsson2021encoding}.
While methods in the field of driving style adaptation primarily focus on ego-vehicle-dependent signals like acceleration and jerk values \cite{choi2021dsagands,lv2019drivingstylebasedco, mohammadnazar2021classifyingtd,khodairy2021drivingbc,kovaceva2020identificationoa,kim2021drivingsc}, the incorporation of the entire driving situation remains elusive.
However, the driving situation has been found to have a significant influence on driving behavior \cite{dong2016characterizing, shouno2018deep, han2019statistical,ghasemzadeh2018utilizing,constantinescu2010driving,chen2019graphical,chen2021driving,hamdar2016weather}.
Furthermore, an alignment between an agent's capability and the given situation increases trust \cite{robert2009individual,petersen2019situational}.
Moreover, individuals' responses to different driving contexts constitute a significant aspect of driving style \cite{chen2019driving}.
Therefore, we propose a situation-aware method to adapt the driving style to the specific human driver.
To fully incorporate the driving situation, we utilize visual feature encoders to learn a representation of the environment.
Building upon this representation, we propose and evaluate two distinct driving style models capable of learning a mapping from the driving situation to the driving behavior, mimicking the specific driver.
\IEEEpubidadjcol  % Formatierung zweite Spalte erste Seite
Our contributions can be summarized as follows:
\begin{enumerate}
    \item A situation-aware driving style adaptation method utilizing learned representations of the driving environment.
    \item An interpretable clustering-based approach for learning situation-dependent driving behaviors and to constrain and control
    the driving style adaptation for specific situations. 
    \item A publicly accessible dataset including \num{1.8} million images and labeled driving behavior data of multiple drivers.
    \item The Entropy-based Cluster Specificity (ECS) metric which uses proxy labels to measure the specificity of the found situation clusters.
    \item The evaluation of unsupervised foundation models (\mbox{DINOv2}) and visual feature encoders pretrained supervised on ImageNet1K for driving style modeling and situation clustering.
    \item The evaluation of MLPs and situation-dependent statistics for driving style modeling and their iterative training capabilities.
\end{enumerate}

\section{Related Work}
In this section, we present an overview of related work in the field of driving style recognition and modeling, highlighting the employed input quantities, the derived output quantities, and the utilized modeling techniques.

\subsection*{Driving Style Input Quantities}
For driving style modeling, the majority of approaches exclusively rely on vehicle BUS time-series data, incorporating information such as acceleration, jerk, and steering wheel angle \cite{choi2021dsagands,lv2019drivingstylebasedco, mohammadnazar2021classifyingtd,khodairy2021drivingbc,kovaceva2020identificationoa,kim2021drivingsc}.
When assessing the driving style solely based on ego-vehicle-centric features, the influence of the driving context is not considered.
However, in various traffic scenarios the driving context either facilitates or constrains decision-making \cite{sagberg2015review}. There is considerable evidence affirming that external conditions significantly influence driving behavior \cite{dong2016characterizing, shouno2018deep, han2019statistical,ghasemzadeh2018utilizing,constantinescu2010driving,chen2019graphical,chen2021driving,hamdar2016weather}.
Although weather has been shown to influence driving behavior significantly \cite{ahmed2018impacts, kilpelainen2007effects, rahman2012analysis, faria2020assessing}, the extent of this variation among individual drivers differs \cite{hamada2016modeling}.
In addition to the influence of weather conditions, traffic also plays a pivotal role, especially when drivers encounter oncoming traffic, leading to deviations from the lane center \cite{rossner2020care,bellem2017can,lex2017objektive,rossner2022also,schlag2015auswirkungen,rosey2009impact,triggs1997effect}.
To incorporate the external context into the driving style analysis, previous works often rely on the isolated inclusion of weather information \cite{hamdar2016weatherar,ahmed2018theio,ghasemzadeh2018utilizingnd,cordero2020recognitionot}, road features \cite{rath2019personalisedlk, ahmed2018theio, hamdar2016weatherar, ghasemzadeh2018utilizingnd, shahverdy2021driverbd, liu2021exploitingmd,cordero2020recognitionot,ghasemzadeh2017driversla}, and traffic data \cite{oezguel2018afu, shahverdy2021driverbd, liu2021exploitingmd, cordero2020recognitionot, chen2021drivingsr, karlsson2021encodinghd}.
Frequently, the relationship with surrounding traffic is extracted from object lists of the vehicle's internal environment perception, as shown in \cite{schrum2023mavericad, zheng2022realtimeds, hajiseyedjavadi2021effectoe, chen2021drivingsr, karlsson2021encodinghd,gao2020personalizedac,moukafih2019aggressivedd,itkonen2020characterisationom}.
In contrast, we utilize raw images from a front-facing camera to fully capture the driving situation without restricting the environment's representation to specific features or scenarios.

\subsection*{Driving Style Output Quantities}
When examining the output quantities, it is evident that the majority of prior methods derive discrete driving style classes \cite{ khodairy2021drivingbc, mohammadnazar2021classifyingtd, xing2020personalizedvt,shahverdy2020driverbd,lv2019drivingstylebasedco}.
While categorizing into broad classes like defensive, moderate, or aggressive provides a high degree of interpretability, defining these classes and their boundaries remains highly subjective.
In contrast, objective model outputs in the form of driving behavior indicators provide an alternative approach \cite{zheng2023towardsdp, li2021combinedtp, wang2020analysisot, yadav2021investigatingte, gao2020personalizedac, itkonen2020characterisationom}.
In addition to these dynamics-oriented indicators, the model parameters of classical mathematical driving behavior models are also predicted \cite{natarajan2022towardad, peralta2022amf, ramezani-khansari2021comparingte, gao2020personalizedac}.
Moreover, scores, such as sportiness or aggressiveness, are derived using predefined calculation procedures \cite{zheng2023towardsdp, tement2022assessmentap, he2022anid, mohammadnazar2021classifyingtd}.
In contrast to the broader driving style classes, the objective indicators of driving behavior offer the advantage of being directly integrable into the personalization of driver assistance systems or automated driving functions through constraints or target variables.
Therefore, we use objective indicators of driving behavior in this work.

\subsection*{Driving Style Modeling Approaches}
On the one hand, driving style modeling often relies on relatively simple rules based on behavioral patterns \cite{tement2022assessmentap, magana2018amf, jardin2020rulebasedds, rossner2020icw, rath2019alk} or statistical models \cite{he2022anid, natarajan2022towardad, cartes2019effectod, hajiseyedjavadi2021effectoe, brück2021investigationop}.
On the other hand, more complex machine-learning-based methods are employed.
This entails utilizing Support Vector Machines (SVMs) \cite{feraud2020asm, savelonas2020classificationod, xue2019rapidds, liu2019researchoc}, K-Nearest Neighbors (KNN) \cite{savelonas2020classificationod, feraud2020asm, bejani2018aca} or Multilayer Perceptrons (MLPs) \cite{savelonas2020hybridtr, feraud2020asm, chen2021semitraj2graphif, jaafer2020dataao} for driving style classification.
Beyond the scope of pure classification, learning-based methods are also applied to learn a driving style and behavior representation \cite{schrum2023mavericad,chen2019driverib} or to predict specific driving-style-related scores \cite{zheng2023towardsdp}.
To capture the temporal aspects of driving behavior and the corresponding driving situation, Recurrent Neural Networks (RNNs) are utilized \cite{schrum2023mavericad, khodairy2021drivingbc, moosavi2021drivingsr, kim2021drivingsc, savelonas2020hybridtr, moukafih2019aggressivedd}.
Even without utilizing images from a vehicle-mounted camera, Convolutional Neural Networks (CNNs) are often employed for driving style modeling \cite{moosavi2021drivingsr, bejani2020convolutionalnn, shahverdy2020driverbd, moukafih2019aggressivedd, liu2021exploitingmd,chen2021semitraj2graphif}.
For converting time-series data of driving measurements into an image-like representation, so-called Driving Operational Pictures (DOPs) are used \cite{dong2016characterizingds, li2019drivingsc, shahverdy2020driverbd, kim2021drivingsc}.

In addition to the frequently utilized supervised approaches for driving style classification or behavior prediction, unsupervised clustering methods are also employed to identify groups of behaviors.
These methods cluster data based on driving behavior metrics such as velocity, accelerations, jerk, or headway values \cite{xing2020personalizedvt, chen2019drivingsc, sun2020ania, zheng2022realtimeds, gao2020personalizedac,lv2019drivingstylebasedco}, or derived representations like risk levels or DOPs embeddings \cite{xue2019rapidds, zhu2020clusteringds}.
This driving behavior clustering is commonly coupled with a preceding reduction of input dimensionality using manifold learning techniques \cite{zheng2022realtimeds, zhu2020clusteringds,shouno2018deepul}.
In contrast, our approach does not rely on clustering behavior data but focuses on clustering the underlying environment representation derived from camera images to model the drivers' individual driving styles in a situation-specific manner.

\section{Datasets}
To assess driving style modeling capabilities of our proposed method, a large dataset with a high scenario diversity is needed to evaluate the situation behavior mapping.
This dataset contains a wide range of situations for pretraining our method and can be considered as fleet data from a manufacturer.
We denote this dataset as the pretrain dataset $\mathcal{D}_P$.
For a fair evaluation of the adaptation capabilities of our method to various drivers and driving situations, data from multiple drivers obtained within similar environmental conditions is needed.
These data represent the behavioral examples of a specific driver collected and used for driving style adaptation in the vehicle.
This dataset is referred to as the validation dataset $\mathcal{D}_V$.

\begin{figure*}[t]
    \centering
   % \frame{
    \includegraphics[page=1,trim={0cm 0 0cm 0}, clip,width=0.9\linewidth]{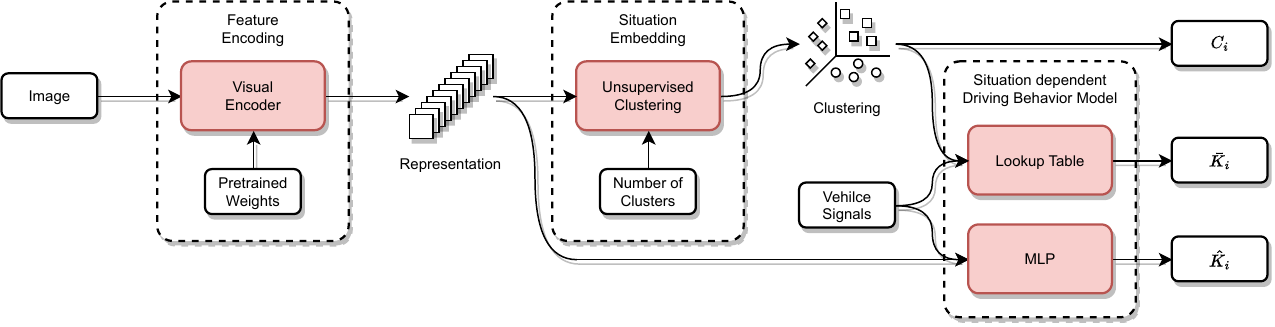}
   % }
    \caption{
        High-level overview of the proposed method.
        Our method consists of a visual feature encoder that infers a representation from an image of a driving situation.
        This encoder is either pretrained on our pretrain dataset, pretrained on \mbox{ImageNet1K}, or a pretrained unsupervised (foundation) model.
        Utilizing this representation, unsupervised clustering is employed to associate each driving situation with a cluster $C_i$.
        This clustering can be used to identify and mask specific driving situations to constrain and control the driving style adaptation.
        We predict the target driving behaviors either with a statistical lookup table that uses the situation cluster $C_i$ for indexing or with MLPs that use the representations from the visual encoder for situation awareness.  
    }
    \label{fig:method}
\end{figure*}

\subsection*{Data Collection}
Since there is a lack of publicly accessible driving datasets covering both image data and driving behavior indicators, as recently discussed in \cite{haselberger2023self}, we collected over 16 hours of driving data from a single test driver using the JUPITER platform \cite{haselberger2022jupiter} as pretrain data.
The data was captured over several months, ensuring a diverse range of road, traffic, and weather conditions.

For the validation data, we utilize the collected driving data from a previously conducted driving style subject study \cite{haselberger2023self} using the same research vehicle as for the pretrain data collection.
Within this study, the driving style of \num{62} subjects was subjectively and objectively analyzed while driving on a given route featuring city, rural, federal, and highway roads.
The data was captured over a small period (two months), ensuring seasonal consistency for the different drivers.
Out of this driver population, we randomly sample five drivers and enrich the dataset with the corresponding captured camera frames.

\subsection*{Dataset Preperation}
To ensure a significant variation of driving situations in the camera stream, the original frame rate of \SI{36}{\hertz} is downsampled to \SI{10}{\hertz}.
Sampling frames randomly to create the training and validation splits likely results in similar driving situations featured in both sets.
To mitigate an overly optimistic evaluation of the generalization ability, we divide the entire driving dataset into equal time segments of three seconds each.
Following this, the segments are randomly assigned to either the training or validation split of $\mathcal{D}_P$ and $\mathcal{D}_V$. 
We use \SI{20}{\percent} of the samples for validation.
To blur vehicle license plates and human faces in the camera frames, we utilize EgoBlur \cite{raina2023egoblur}. 
Furthermore, all subject-related data, including the socio-demographics, are anonymized. 

\subsection*{Dataset}
The final dataset is composed as follows: the pretrain set $\mathcal{D}_P$ is split into a training subset $\mathcal{D}_{P,T}$ with \num{242887} samples, and a validation subset $\mathcal{D}_{P,V}$ with \num{61400} samples.
Similarly, the validation set $\mathcal{D}_V$ is split into a training subset $\mathcal{D}_{V,T}$ and a validation subset $\mathcal{D}_{V,V}$ with \num{138572} and \num{34767} samples.
Each subset consists of $1280 \times 960$ images, driving behavior indicators like the distance to the lane center or longitudinal headway distances, vehicle signals like velocity or accelerations, as well as traffic conditions and road type labels.
The entire unfiltered pretrain data and the unfiltered validation data of the five drivers (\num{1.8} million samples), as well as the processed datasets, are publicly available at \href{https://huggingface.co/datasets/jHaselberger/SADC-Situation-Awareness-for-Driver-Centric-Driving-Style-Adaptation}{\textbf{huggingface.co/datasets/jHaselberger/SADC-Situation-Awareness-for-Driver-Centric-Driving-Style-Adaptation}} under the \mbox{CC\,BY\,4.0\,DEED} license.

% \section{Situation Aware Driving Behavior}
\section{Method}
Our proposed method consists of three components: visual feature encoding, situation embedding, and situation-dependent driving behavior modeling.
A graphical overview is provided in \autoref{fig:method}.
Without the loss of generalization, we select the distance to the center lane ($d_{\mathrm{CL}}$) as the target variable to characterize the lateral driving style.
Previous studies show that this quantity is highly driver-heterogenous and can be integrated into the development and evaluation of lateral driving functions \cite{barendswaard2019acm, haselberger2023exploring, haselberger2023self, hofer2020attribute}.

\subsection*{Visual Feature Encoding}
To get a representation $R_i$ of a given driving situation $S_i$, we pretrain a visual feature encoder $E(S_i)$ on our pretrain dataset $\mathcal{D}_{P,T}$.
As the loss function $\mathcal{L}$, we calculate the mean squared error (MSE) between the predicted ($\hat{d}_{\mathrm{CL}}$) and the measured distance to the center lane of the human driver ($d_{\mathrm{CL}}$) for the given situation:
\begin{equation}
    \mathcal{L} = \frac{1}{N} \sum_{i=1}^{N}(d_{\mathrm{CL}}-\hat{d}_{\mathrm{CL}})^2
\label{eq:featureEncoderLossFunction}
\end{equation}
Furthermore, we experiment with visual feature encoders pretrained supervised on ImageNet1K \cite{russakovsky2015imagenet} and unsupervised on curated data to evaluate the performance of behavior prediction and situation clustering based on representations obtained from off-the-shelf encoders.

\subsection*{Situation Embedding}
Given the multitude of diverse road, weather, and traffic situations encountered in real-world driving, the underlying situation space is not easily definable and manageable using traditional rule-based approaches.
Therefore, we employ unsupervised clustering to associate each driving situation $S_i$ with a cluster $C_i$ utilizing the representation $R_i = E(S_i)$.
In this way, we model the drivers' individual driving styles in a situation-specific manner without prior knowledge of the situation space.
Moreover, besides a low computation effort, clustering provides high interpretability.
The identified clusters can be examined utilizing the given mapping from the situation embeddings to the camera images and the corresponding vehicle signals.
In this work, we use different variants of k-means clustering with the target number of clusters $N_C$ as an adjustable parameter to regulate the situation-specificness for driving style adaption. 

\subsection*{Situation Aware Driving Behavior}
% \subsubsection*{Situation-Dependent Driving Behavior Modeling}
Using the assigned situation cluster $C_i$, we predict the target driving behavior indicators $K_i$ with a statistical lookup table.
To train each of the $N_C$ entries of the lookup table, we gather objective driving behavior samples for each assigned situation embedding and calculate the target behavior indicators $\bar{K_i}$ based on derived statistics of the $N_{C_i}^{d_{\mathrm{CL}}}$ collected driving samples $d_{\mathrm{CL}}$:
\begin{equation}
    \bar{K_i} = \frac{1}{N_{C_i}^{d_{\mathrm{CL}}}}\sum_{j=1}^{N_{C_i}^{d_{\mathrm{CL}}}}d_{\mathrm{CL},j}
\label{eq:ki}
\end{equation}
Based on the possible large amount of different situation clusters $N_C$, this is an efficient statistic-based method to predict the driving behavior indicators.

% \subsubsection*{End-to-End Driving Behavior Modeling}
To further compare our cluster-based approach, we also train driving behavior models end-to-end directly on the situation images $\hat{K}_i = H(E(S_i))$, where $H$ refers to fully-connected layers to obtain the final prediction $\hat{K}_i$.
For a direct comparison to situation-dependent driving behavior modeling, we use the same visual feature encoder architectures.
In contrast to the cluster-based approach that explicitly decouples the driving situation and the behavior modeling, the end-to-end approach only implicitly considers the driving situation, which reduces interpretability.
From the perspective of a manufacturer, explicit situation-behavior mapping provides the possibility to constrain and control the driving style adaptation for specific situations.

\subsection*{Driver-Centric Driving Style Adaptation}
Given substantial evidence that every driver has their unique driving style \cite{dong2016characterizing, woo2018dynamic,brambilla2017comparison,sun2020intention,lin2014overview,tement2022assessment,kim2021driving}, we adapt our model towards the driving style of specific drivers. 
Therefore, we freeze the visual feature encoder and the clusters learned on the pretrain dataset $\mathcal{D}_{P,T}$.
Only the entries of the situation-dependent lookup table are updated using the driver-specific behavior data $\mathcal{D}_{V,T}$.
As a second approach, we train fully-connected predictor heads on the representations of the frozen visual feature encoder for each specific driver separately.
Separating the training of the visual encoder and clustering from behavior modeling allows training these two components on a wide variety of situations obtained from fleet data not necessarily encountered by a single driver.
Furthermore, this split enables the training of time, data, and resource-consuming feature encoders by the manufacturer on dedicated computation machines rather than on the actual vehicles.
Similarly, the pretraining of the clusters provides the possibility to share a common situation-behavior-mapping across all vehicles, facilitating consistency and testability from the manufacturer's perspective.
On top of this, clustering can mitigate the effects of catastrophic forgetting when adapting to new situations.
The driver-centric training of the situation-dependent lookup table and fully-connected heads can be done directly on the vehicle. 
% Furthermore, clustering is a simple approach for fast learning of personalized driving styles.

\subsection*{Integration into ADAS / HAF}
Compared to direct control quantities like steering angle or gas pedal position, the derived driving behavior indicators from our model can be treated as constraints or target values for low-level controllers like in \cite{cartes2019effectod, karlsson2021encodinghd, bae2020selfdrivingla, gao2020personalizedac}.
Decoupling driving behavior indicators from the actual control quantities ensures a driving style adaptation safeguarded by the low-level controller.
Moreover, our method is not restricted to lateral indicators such as the distance to the lane center and, in theory, can be generalized to other use cases, such as adapting longitudinal headway distances for Adaptive Cruise Control (ACC).
Besides using the clustering as indexing for the driving behavior lookup table, the situation embeddings can also be seen as additional output of our method.
This output can be further used to mask specific situations for other driving behavior models, like the MLPs used in this work.
Decoupling situation clustering from driving behavior modeling provides the possibility of employing various types of visual feature encoders for both tasks.
\section{Experiments}
We conduct various experiments to evaluate our method regarding its capabilities to model the human situation-aware driving behavior, its adaptability to different drivers, and the specificity of the identified situation clusters.
% Across all experiments, we compare visual feature encoders trained supervised on our dataset, pretrained on ImageNet1K, and pretrained unsupervised on curated data from different sources.
% Furthermore, for driving behavior modeling we compare end-to-end trained linear and non-linear fully-connected predictors as well as cluster-based situation-dependent statistics.
For all experiments, we report mean and standard deviations across five runs.

\subsection*{Metrics}
For evaluation of the lateral driving behavior modeling, we utilize the root-mean-square error (RMSE) between the human and the predicted distance to the lane center $\hat{d}_{\mathrm{CL}}$:
\begin{equation}
    \mathrm{RMSE} = \sqrt{\frac{1}{N} \sum_{i=1}^{N}(d_{\mathrm{CL}}-\hat{d}_{\mathrm{CL}})^2}
\label{eq:rmse}
\end{equation}

For assessing the adaptation performance on the validation subset $\mathcal{D}_{V,V}$, we average the error across all five drivers.
We report RMSE values for the entire validation datasets $\mathcal{D}_{P,V}$ and $\mathcal{D}_{V,V}$ (All) and on  subsets containing only rural situations (Rural Only).
As additional benchmarks, we refer to the curve-cutting-gradient-based driving styles from \cite{haselberger2023exploring} without considering the driving situation.
These static driving styles consist of constant lane centering (Rail), minimal curve cutting (Passive), and a sportive driving style with high curve cutting gradients (Sportive).
The statistical significance of the mean differences between our proposed method and the static driving styles was analyzed using jamovi \cite{jamovi2023jamovi}, an open-source statistical software.

To quantitatively evaluate the clustering of the representations into specific situations, we propose the Entropy-based Cluster Specificity (ECS) metric.
As the underlying situation space is unknown and cannot be clearly described, our metric incorporates $N_L$ discrete labels, which act as proxy labels for the driving situation.
In our case, we define six proxy labels: road type, curvature, as well as type and distance of oncoming and leading vehicles.
Thereby, the l-th label is binned into $N_{B_l}$ bins. 
Using the learned mapping of the driving situation to the c-th cluster centroid, we can select a subset $L_{c}$ of all label data $L$.
For each label $L_{c,l}$ in the selected subset, we utilize the normalized Shannon entropy \cite{wilcox1967indices}:
\begin{equation}
    h(L_{c,l}) = -\frac{\sum_{i=1}^{N_{B_l}}p(L_{c,l,i}) \log p(L_{c,l,i})}{\log(N_{B_l})}
\label{eq:entropy}
\end{equation}
to define the specificity value $s(L_{c,l}) = 1 - h(L_{c,l})$.
We employ the inverse of the entropy as we want to identify highly specialized clusters.
We then combine the centroid-wise maximum and average specificity values:
\begin{equation}
    \mathrm{ECS} = \frac{1}{N_C}\sum_{c=1}^{N_C} \left( \max_{l \in N_L}s(L_{c,l}) \times \frac{1}{N_L} \sum_{l=1}^{N_L}s(L_{c,l}) \right)
\label{eq:ecs}
\end{equation}
We balance contributions from highly specialized centroids by taking the maximum and contributions from centroids specialized across multiple labels by calculating the average over all $N_L$ labels.
For the final ECS score, we calculate the average across all clusters $N_C$.
The ECS metric is bound between \num{0} and \num{1}, as it is derived from the normalized Shannon entropy.

\subsection*{Models}
For our visual feature encoder, we experiment with the convolution-based \mbox{ResNet-18} \cite{he2016deep}, \mbox{ResNet-50} \cite{he2016deep}, \mbox{ResNeXt-50} \cite{xie2017aggregated} models, and a large attention-based visual image transformer \mbox{(ViT-L)} \cite{dosovitskiy2020image}. 
We either pretrain these models on our dataset or use their pretrained versions on \mbox{ImageNet1K} \cite{russakovsky2015imagenet}.
As an unsupervised foundation model for the visual feature encoder, we select DINOv2 with registers \cite{darcet2023vision} in the sizes small \mbox{(DINO-S)}, big \mbox{(DINO-B)}, large \mbox{(DINO-L)}, and giant \mbox{(DINO-G)}.
All \mbox{DINOv2} models are based on visual image transformers.
Visual, unsupervised foundation models like Dinov2 are intended to learn representations that can directly be used for any image-level or pixel-level task.
For clustering of the representations, we utilize classical and spherical unsupervised K-Means Clustering \cite{macqueen1967classification,johnson2019billion}.
As predictor heads for the driving behavior based on the representations, we experiment with fully-connected linear layers and MLPs.

\subsection*{Implementation Details}
We implement all methods in PyTorch 2.1.1 and train them on a single machine with up to eight NVIDIA A100 GPUs.
For GPU accelerated training of both K-Means variants, we utilize the Faiss library \cite{johnson2019billion}.
For training of the feature encoders, we resize the input images to height \num{224}, crop $224 \times 224$ patches with center cropping, apply AugMix augmentations \cite{hendrycks2019augmix} on the images, and normalize them with mean \num{0.5} and standard deviation \num{0.5}.
We use AdamW \cite{loshchilov2017decoupled} with standard parameters as optimizer, a cosine annealing learning rate schedule, a batch size of \num{256}, and tune learning rates as well as epochs separately for each model. 
For the supervised pretrained models, we utilize the weights provided by torchvision 0.16.1 \cite{torchvision2016}.
Since there are different versions of ImageNet1K weights, we choose the weights with the best reported performance on ImageNet1K.
We follow the original implementation of Dinov2 and use the provided weights \cite{darcet2023vision} to infer representations of our dataset.
All MLP heads consist of three layers with $[2048,2048,1]$ units, ReLU activations, batch normalization \cite{ioffe2015batch}, and a tanh output activation.
Before further processing by K-Means clustering or the fully-connected heads, we standardize the representations by removing the mean and scaling to unit variance.
Our implementation is publicly available at \href{https://github.com/jHaselberger/SADC-Situation-Awareness-for-Driver-Centric-Driving-Style-Adaptation}{\textbf{github.com/jHaselberger/SADC-Situation-Awareness-for-Driver-Centric-Driving-Style-Adaptation}}.

\subsection*{Situation Aware Driving Behavior}
To test the capabilities of our models to predict the human situation-aware driving behavior, we train neural-network-based (NN) behavior predictors end-to-end on the pretrain dataset $\mathcal{D}_{P,T}$ and report the RMSE results on $\mathcal{D}_{P,V}$ in \autoref{tab:supervisedPretrain}.
We use the best-performing feature encoders of the end-to-end training for driving situation clustering (DSC). 
For training the driving situation dependent statistics (DSDS), the number of clusters $N_C$ is varied from \num{5} to \num{3000} and the best-performing configuration is reported.
Compared to the static driving styles, both NN and DSC predict human driving behavior with significantly (\pValuePL{}) lower mean errors according to the Post-hoc test of a robust analysis of variance (ANOVA) \anovaPL{\robustANOVAName}{2.0}{48990}{15814}{.001}.  
Overall, the end-to-end trained models lead to the lowest RMSE values for both domains.
However, the DSC approach leads to more stable results, indicated by the lower standard deviations.
For the end-to-end method, \mbox{ResNet-18} performes the best in our experiments with an RMSE value of $0.0806 \pm 0.0014$.
However, as we observe in the results of the DSC method, the larger representation sizes of the \mbox{ResNet-50} and \mbox{ResNeXt-50} encoders lead to performance improvements when the behavior prediction is decoupled from training the feature encoder.
Furthermore, it can be seen that the classical K-Means variant leads to better results compared to the spherical counterpart.
It is evident that, unlike static driving style models, our learning-based methods deliver slightly better results in all situations compared to the rural subset.
On the one side, this may be attributed to the higher amount of available training data. On the other side, the rural-only subset consists of a higher behavior variability, given the higher variance in curve radii.
\begingroup

\setlength{\tabcolsep}{2pt} % Default value: 6pt

\begin{table}[]
    \caption{
        Results of our methods on $\mathcal{D}_{P,V}$ with visual feature encoders pretrained on $\mathcal{D}_{P,T}$.
    }
    \centering
    \scriptsize
    \begin{tabular}{lcclclc}
    \toprule
    \textbf{}                                  & \textbf{Visual}  & \textbf{Behavior}  & \textbf{} & \multicolumn{1}{c}{\textbf{All}}                           & \textbf{} & \multicolumn{1}{c}{\textbf{Rural Only}}                         \\ \cline{5-5} \cline{7-7}\clineSpacing
    \textbf{}                                  & \textbf{Encoder} & \textbf{Predictor} & \textbf{}  & \textbf{RMSE}  & \textbf{}  & \textbf{RMSE}  \\
    \midrule
    \multirow{3}{*}{\rotatebox{90}{NN}}        & ResNet-18         & MLP                &           & \textbf{0.0806 $\pm$ 0.0014}     &           & \textbf{0.0923 $\pm$ 0.0022}    \\
                                               & ResNet-50         & MLP                &           & 0.0822 $\pm$ 0.0013              &           & 0.0978 $\pm$ 0.0013             \\
                                               & ResNeXt-50        & MLP                &           & 0.0823 $\pm$ 0.0013              &           & 0.0984 $\pm$ 0.0019             \\
    \midrule
    \midrule
                                % &                  &                    &           &                                  &                                  &           &                                 &                                 \\
    \multirow{6}{*}{\rotatebox{90}{DSC}}       & ResNet-18         & DSDS-KM            &           & 0.1075 $\pm$ 0.0001              &           & \underline{0.1080 $\pm$ 0.0003} \\
                                               &                  & DSDS-KMS            &           & 0.1159 $\pm$ 0.0005              &           & 0.1170 $\pm$ 0.0004 \\
                                               & ResNet-50         & DSDS-KM            &           & 0.1035 $\pm$ 0.0005              &           & 0.1314 $\pm$ 0.0005             \\
                                               &                  & DSDS-KMS            &           & 0.1093 $\pm$ 0.0004              &           & 0.1332 $\pm$ 0.0004             \\
                                               & ResNext-50        & DSDS-KM            &           & \underline{0.1023 $\pm$ 0.0008}  &           & 0.1272 $\pm$ 0.0009             \\
                                               &                  & DSDS-KMS            &           & 0.1077 $\pm$ 0.0007              &           & 0.1308 $\pm$ 0.0005             \\
    \midrule
    \midrule
                                % &                  &                    &           &                                  &                                  &           &                                 &                                 \\
    \multirow{3}{*}{\rotatebox{90}{Static}}     &  & Passive            &           & 0.2519                           &           & 0.2115                          \\
                                &                  & Rail               &           & \underline{0.2314}               &           & \underline{0.2027}              \\
                                &                  & Sportive           &           & 0.2801                           &           & 0.2460                          \\
    \bottomrule  
    \end{tabular}
    \label{tab:supervisedPretrain}
    \end{table}

\endgroup

\subsection*{Driver-Centric Driving Style Adaptation}
Since the previous experiment demonstrates the general modeling capabilities of our method, we further investigate the adaptability to different drivers.
Therefore, we freeze the feature encoders and the situation clustering pretrained on $\mathcal{D}_{P,T}$ and train the predictor heads for each driver in the dataset $\mathcal{D}_{V,T}$ separately.
As shown in \autoref{tab:supervisedVal} and by the Post-hoc tests of a robust ANOVA \anovaPL{\robustANOVAName}{2.0}{5755}{2650}{.001}, the learning-based methods outperform the static driving styles significantly with \pValuePL{}.
For the RMSE metric, the MLP behavior predictor performs the best, followed by DSDS and the linear model.
Similar to the pretrain experiments, DSDS turned out to be the most stable model.
As indicated by the DSC results in \autoref{tab:supervisedPretrain}, a larger representation size positively impacts performance in most cases when the predictor heads for the different drivers are trained separately from the visual feature encoder.
Moreover, the lower RMSE values on $\mathcal{D}_{V,V}$ compared to $\mathcal{D}_{P,V}$ show that the feature encoders pretrained on $\mathcal{D}_{P,T}$ provide beneficial representations for situation-dependent driving behavior modeling of different drivers.
This can also be seen in the reduced performance gap between the two domains.
These results support the underlying concept of our adaptation method to decouple training of the visual feature encoder from behavior prediction.
This enables the incorporation of a wide variety of situations obtained from fleet data and to share a common situation behavior mapping.
\begingroup

\setlength{\tabcolsep}{2pt} % Default value: 6pt

\begin{table}[]
    \caption{
        Results of our methods on $\mathcal{D}_{V,V}$ with visual feature encoders pretrained on $\mathcal{D}_{P,T}$ and prediction heads trained on $\mathcal{D}_{V,T}$.
    }
    \centering
    \scriptsize
    \begin{tabular}{lcclclc}
    \toprule
    \textbf{}                   & \textbf{Visual}  & \textbf{Behavior}  & \textbf{} & \multicolumn{1}{c}{\textbf{All}}                           & \textbf{} & \multicolumn{1}{c}{\textbf{Rural}}                         \\ \cline{5-5} \cline{7-7}\clineSpacing
    \textbf{}                   & \textbf{Encoder} & \textbf{Predictor} & \textbf{}   & \textbf{RMSE}  & \textbf{}   & \textbf{RMSE} \\
    \midrule
    \multirow{6}{*}{\rotatebox{90}{NN}} & ResNet-18 & MLP                &           & 0.0737 $\pm$ 0.0010               &           & 0.0752 $\pm$ 0.0021                \\
                                &                   & Linear             &           & 0.1750 $\pm$ 0.0048               &           & 0.1809 $\pm$ 0.0083                \\
                                & ResNet-50         & MLP                &           & 0.0685 $\pm$ 0.0012               &           & 0.0755 $\pm$ 0.0009                \\
                                &                   & Linear             &           & 0.1570 $\pm$ 0.0028               &           & 0.1506 $\pm$ 0.0040                \\
                                & ResNeXt-50        & MLP                &           & \textbf{0.0677 $\pm$ 0.0010}      &           & \textbf{0.0739 $\pm$ 0.0014}       \\
                                &                   & Linear             &           & 0.1566 $\pm$ 0.0036               &           & 0.1551 $\pm$ 0.0034                \\
                                \midrule
                                \midrule
                                                            % &                  &                    &           &                                  &                                  &           &                                 &                                 \\
                                \multirow{6}{*}{\rotatebox{90}{DSC}}       & ResNet-18         & DSDS-KM            &           & 0.1027 $\pm$ 0.0005              &           & \underline{0.0954 $\pm$ 0.0009} \\
                                                                           &                   & DSDS-KMS           &           & 0.1115 $\pm$ 0.0009              &           & 0.1086 $\pm$ 0.0009 \\
                                                                           & ResNet-50         & DSDS-KM            &           & 0.1026 $\pm$ 0.0011              &           & 0.1084 $\pm$ 0.0013             \\
                                                                           &                   & DSDS-KMS           &           & 0.1087 $\pm$ 0.0011              &           & 0.1096 $\pm$ 0.0011             \\
                                                                           & ResNext-50        & DSDS-KM            &           & \underline{0.1006 $\pm$ 0.0004}  &           & 0.1149 $\pm$ 0.0009             \\
                                                                           &                   & DSDS-KMS           &           & 0.1053 $\pm$ 0.0007              &           & 0.1158 $\pm$ 0.0013             \\
                                \midrule
                                \midrule
                                % &                  &                    &           &                                   &                                   &           &                                  &                                   \\
    \multirow{3}{*}{\rotatebox{90}{Static}}    &   & Passive            &           & \underline{0.2653}                &           & \underline{0.2383}                \\
                                &                  & Rail               &           & 0.2716                            &           & 0.2453                            \\
                                &                  & Sportive           &           & 0.2738                            &           & 0.2470                            \\
    \bottomrule
    \end{tabular}
    \label{tab:supervisedVal}
    \end{table}

\endgroup

\subsection*{Impact of Clusters Quantity}
\begin{figure*}[t]
    \centering
    \small
    \begin{tabular}{ccc}
        \includegraphics[trim={-0.2cm 0 -0.9cm -0cm}, clip, width=.3\linewidth,valign=m]{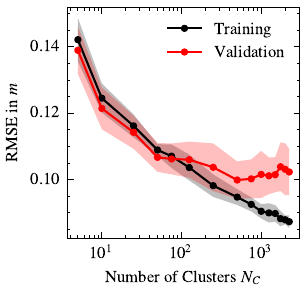} & 
        \includegraphics[trim={0cm 0 -0.95cm 0}, clip,width=.3\linewidth,valign=m]{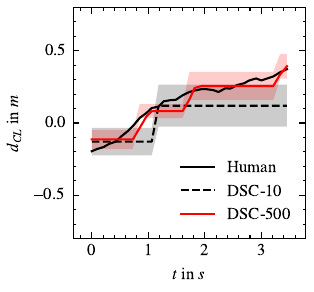} & 
        \includegraphics[trim={0cm 0 -0.95cm 0}, clip,width=.3\linewidth,valign=m]{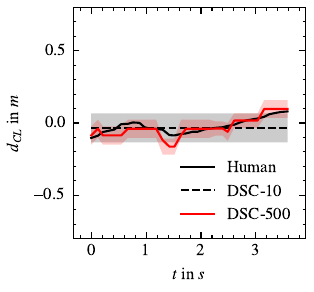} \\ 
        a) & b) & c) \\
    \end{tabular}
    \caption{
        a) Training and validation RMSE of the DSC method on $\mathcal{D}_{V,V}$ for an increasing number of clusters $N_C$ utilizing the  ResNeXt-50 feature encoder pretrained on $\mathcal{D}_{P,T}$.
        b) and c) Predictions of the DSC approach with ResNeXt-50 feature encoding for two specific driving situations with $N_C = 10$ (DSC-10) and $N_C = 500$ (DSC-500). 
    }
\label{fig:nCLusterWS}
\end{figure*}

To study the impact of the number of clusters, we vary the cluster quantity $N_C$ from \num{5} up to \num{3000} while keeping the remaining behavior modeling the same.
As shown in \autoref{fig:nCLusterWS} a), a decreasing trend in the resulting RMSE values can be observed during training across all drivers.
However, for a higher number of clusters, the validation curve shows convergence or slight overfitting behavior.
This confirms the dependency of the driving behavior modeling accuracy concerning $N_C$.
As shown in \autoref{fig:nCLusterWS} b) and c), a lower number of clusters results in a coarser estimate of the driving behavior while maintaining the general trend in curve cutting.
This can be attributed to the higher number of driving samples assigned to the same situation cluster, which are taken into account for the statistic-based driving style modeling.
Increasing the number of clusters up to the optimum leads to a higher level of specialization of the learned clusters and a more situation-dependent capture of the human driving behavior, resulting in a more accurate reproduction of the human driving style.

\subsection*{Impact of Pretrained Visual Feature Encoders}
\begin{figure*}[h!]
    \centering
    \small
    \vspace{1cm}
    \begin{tabular}{ccc}
        \textbf{Feature Encoder Pretrained on} & \textbf{Feature Encoder Pretrained on} & \textbf{Feature Encoder Pretrained on} \\
        \textbf{$\mathcal{D}_{P}$ (ours)} & \textbf{ImageNet1K} & \textbf{Currated Data (Dinov2)} \\
        ~ & ~ & ~ \\
        \includegraphics[trim={0cm 0 0cm 0}, clip, width=.3\linewidth,valign=m]{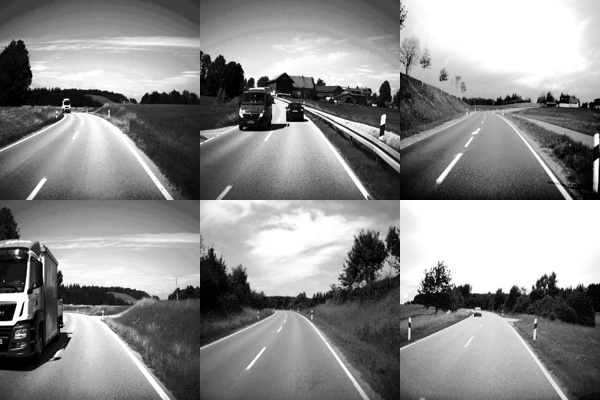} & 
        \includegraphics[trim={0cm 0 0cm 0}, clip,width=.3\linewidth,valign=m]{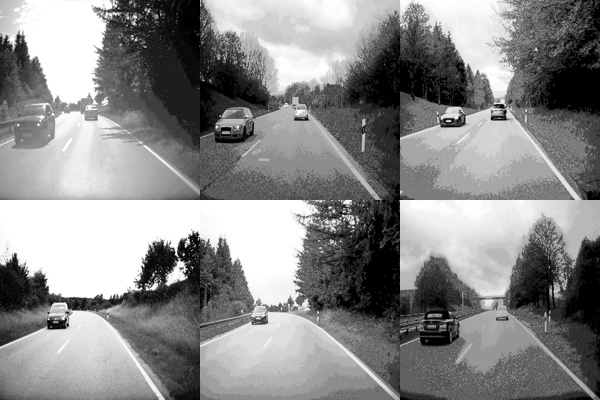} & 
        \includegraphics[trim={0cm 0 0cm 0}, clip,width=.3\linewidth,valign=m]{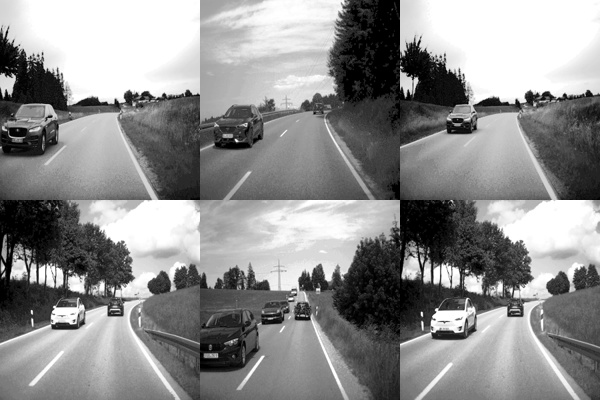} \\ 
        ~ & ~ & ~ \\
        \includegraphics[trim={0cm 0 0cm 0}, clip, width=.3\linewidth,valign=m]{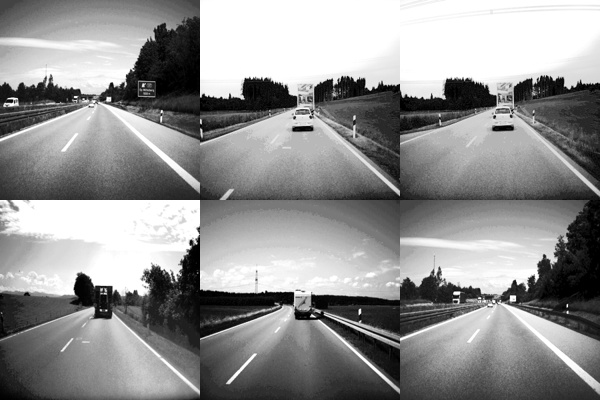} & 
        \includegraphics[trim={0cm 0 0cm 0}, clip,width=.3\linewidth,valign=m]{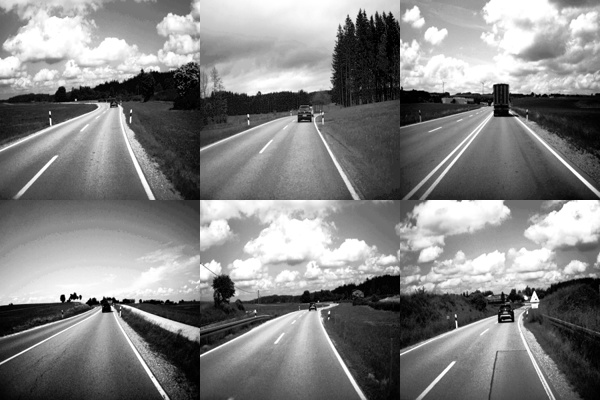} & 
        \includegraphics[trim={0cm 0 0cm 0}, clip,width=.3\linewidth,valign=m]{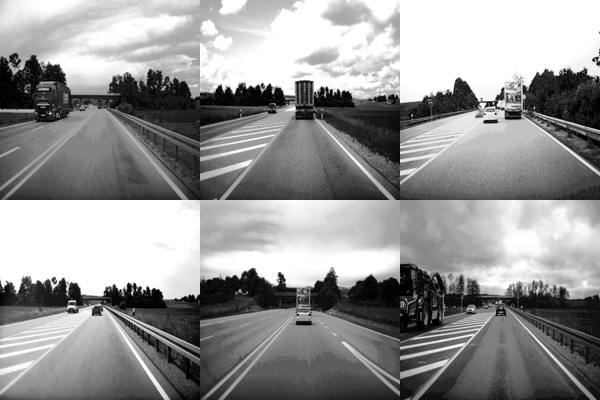} \\ 
        ~ & ~ & ~ \\
        \includegraphics[trim={0cm 0 0cm 0}, clip, width=.3\linewidth,valign=m]{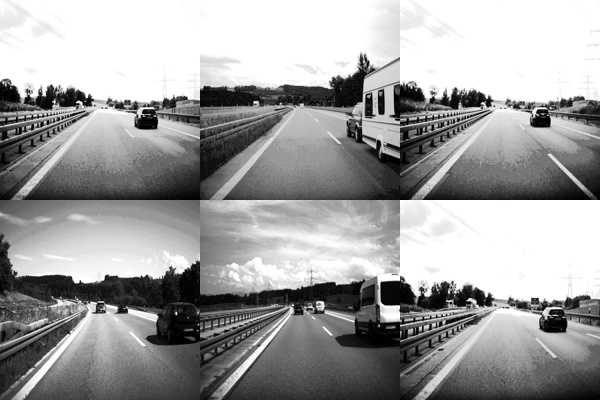} & 
        \includegraphics[trim={0cm 0 0cm 0}, clip,width=.3\linewidth,valign=m]{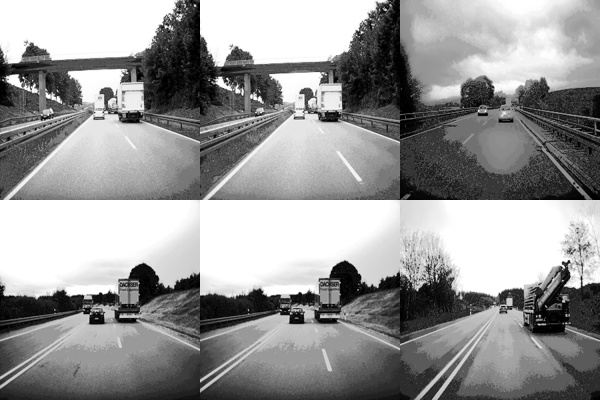} & 
        \includegraphics[trim={0cm 0 0cm 0}, clip,width=.3\linewidth,valign=m]{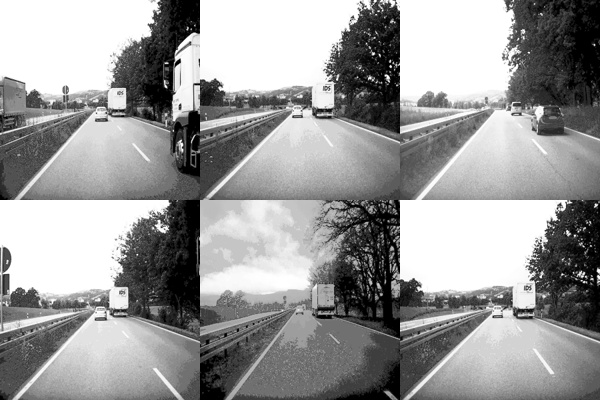} \\ 
        ~ & ~ & ~ \\
        \includegraphics[trim={0cm 0 0cm 0}, clip, width=.3\linewidth,valign=m]{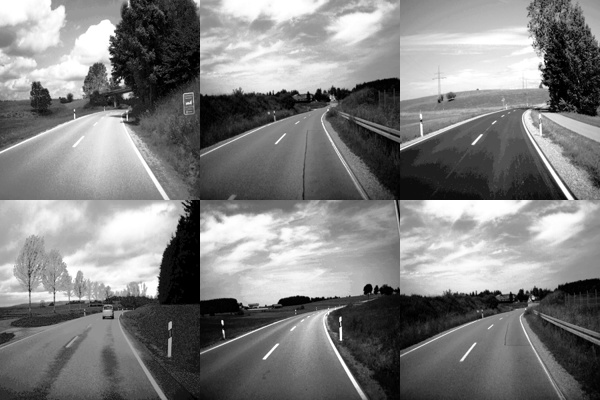} & 
        \includegraphics[trim={0cm 0 0cm 0}, clip,width=.3\linewidth,valign=m]{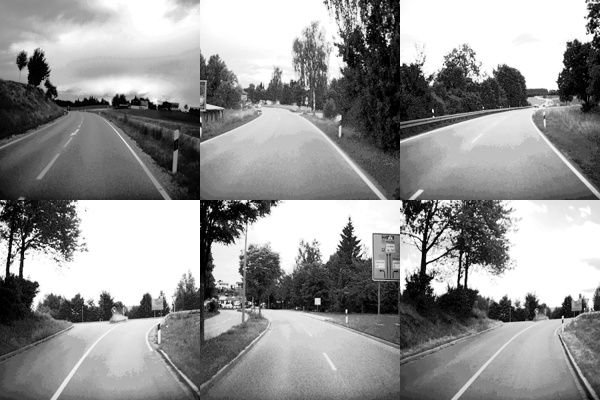} & 
        \includegraphics[trim={0cm 0 0cm 0}, clip,width=.3\linewidth,valign=m]{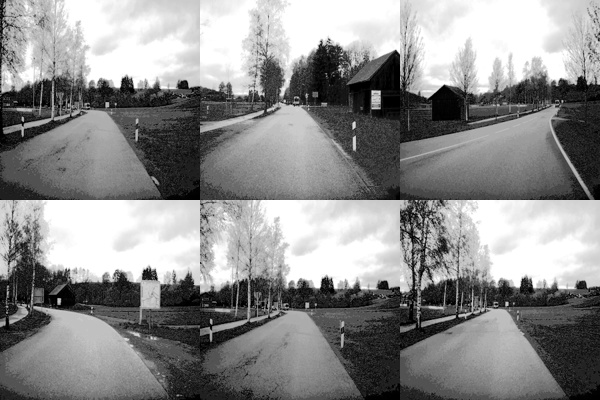} \\ 
        ~ & ~ & ~ \\
        \includegraphics[trim={0cm 0 0cm 0}, clip, width=.3\linewidth,valign=m]{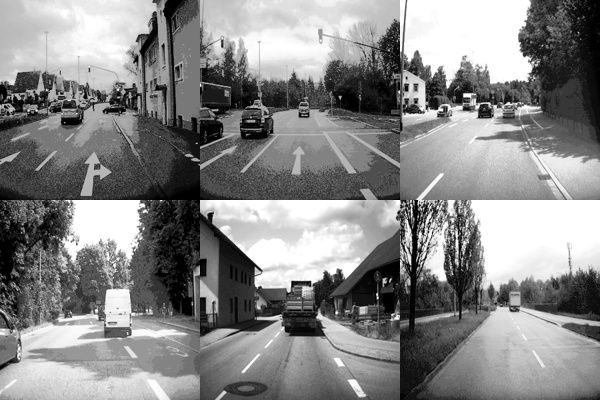} & 
        \includegraphics[trim={0cm 0 0cm 0}, clip,width=.3\linewidth,valign=m]{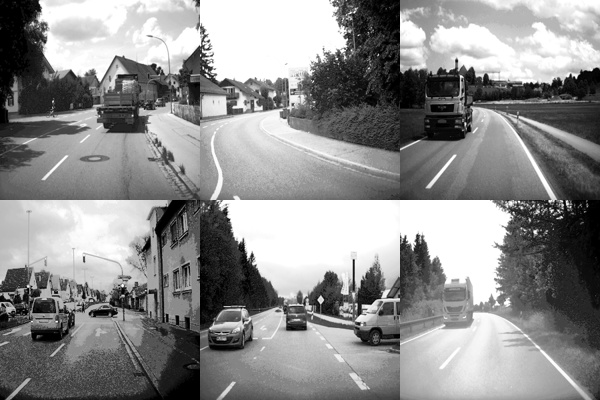} & 
        \includegraphics[trim={0cm 0 0cm 0}, clip,width=.3\linewidth,valign=m]{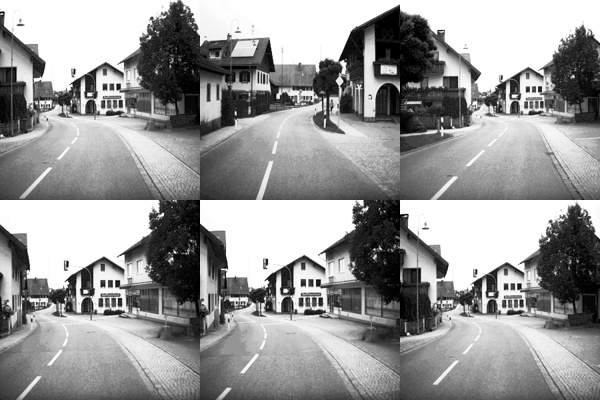} \\ 
    \end{tabular}
    \caption{
        Sample images of learned situation clusters using the representations from the visual feature encoders pretrained on our pretrain dataset $\mathcal{D}_{P,T}$, ImageNet1K, and in an unsupervised manner on curated data from different sources.
        For each situation cluster, we sample six images randomly from the set of assigned driving situations of $\mathcal{D}_{V,T}$.
        In the first four rows, we aim to highlight various aspects of potential driving situations, including oncoming traffic, following vehicles, overtaking, and driving on rural roads.
        In the last row, possible shortcomings of the clusters, such as unclear driving situations or over-specification, are shown.
    }
  \label{fig:clusterImages}
  \end{figure*}

\begingroup

\setlength{\tabcolsep}{2pt} % Default value: 6pt

\begin{table*}[]
    \caption{
        Results of our methods on $\mathcal{D}_{V,V}$ with visual feature encoders pretrained supervised on ImageNet1K (IN) or unsupervised on currated data (Dino) and prediction heads trained on $\mathcal{D}_{V,T}$.
    }
    \centering
    \scriptsize
    \begin{tabular}{lcclclcp{4pt}cclclc}
    \toprule
    \textbf{}                   & \textbf{Visual}  & \textbf{Behavior}  & \textbf{} & \multicolumn{1}{c}{\textbf{All}}                           & \textbf{} & \multicolumn{1}{c}{\textbf{Rural Only}} & & \textbf{Visual}  & \textbf{Behavior}  & \textbf{} & \multicolumn{1}{c}{\textbf{All}}                           & \textbf{} & \multicolumn{1}{c}{\textbf{Rural Only}}    \\ \cline{5-5}\cline{7-7}\cline{12-12}\cline{14-14}\clineSpacing 
    \textbf{}                   & \textbf{Encoder} & \textbf{Predictor} & \textbf{}   & \textbf{RMSE}  & \textbf{}   & \textbf{RMSE} & & \textbf{Encoder} & \textbf{Predictor} & \textbf{}   & \textbf{RMSE}  & \textbf{}   & \textbf{RMSE} \\
    \midrule
    \multirow{8}{*}{\rotatebox{90}{NN}}   & ResNet18-IN            & MLP                &           & \textbf{0.1652 $\pm$ 0.0011}      &           & \textbf{0.1554 $\pm$ 0.0008}   &       & Dino-S  & MLP       &       & 0.1658 $\pm$ 0.0007              &       & 0.1654 $\pm$ 0.0017               \\
                                          &                        & Linear             &           & 0.3845 $\pm$ 0.0038               &           & 0.3043 $\pm$ 0.0013            &       &         & Linear    &       & 0.4823 $\pm$ 0.0059              &       & 0.3255 $\pm$ 0.0038               \\
                                          & ResNet50-IN            & MLP                &           & 0.1731 $\pm$ 0.0004               &           & 0.1614 $\pm$ 0.0008            &       & Dino-B  & MLP       &       & 0.1665 $\pm$ 0.0010              &       & 0.1610 $\pm$ 0.0017               \\
                                          &                        & Linear             &           & 0.4513 $\pm$ 0.0013               &           & 0.2801 $\pm$ 0.0014            &       &         & Linear    &       & 0.4903 $\pm$ 0.0034              &       & 0.3086 $\pm$ 0.0035               \\
                                          & ResNeXt50-IN           & MLP                &           & 0.1732 $\pm$ 0.0008               &           & 0.1609 $\pm$ 0.0007            &       & Dino-L  & MLP       &       & 0.1684 $\pm$ 0.0005              &       & 0.1612 $\pm$ 0.0008               \\
                                          &                        & Linear             &           & 0.4627 $\pm$ 0.0014               &           & 0.2700 $\pm$ 0.0012            &       &         & Linear    &       & 0.4842 $\pm$ 0.0037              &       & 0.2946 $\pm$ 0.0077               \\
                                          & ViT-L-IN               & MLP                &           & 0.1752 $\pm$ 0.0011               &           & 0.1721 $\pm$ 0.0011            &       & Dino-G  & MLP       &       & \textbf{0.1653 $\pm$ 0.0001}     &       & \textbf{0.1597 $\pm$ 0.0021}      \\
                                          &                        & Linear             &           & 0.4118 $\pm$ 0.0077               &           & 0.3115 $\pm$ 0.0019            &       &         & Linear    &       & 0.4849 $\pm$ 0.0048              &       & 0.2754 $\pm$ 0.0012               \\
    \midrule
    \midrule
                                % &                  &                    &           &                                  &                                  &           &                                 &                                 \\
    \multirow{8}{*}{\rotatebox{90}{DSC}}       & ResNet18-IN          & DSDS-KM            &           & \underline{0.2366 $\pm$ 0.0008}&           & \underline{0.2151 $\pm$ 0.0014}&       & Dino-S  & DSDS-KM     &       & 0.2289 $\pm$ 0.0023                  &       & 0.2102 $\pm$ 0.0048             \\
                                               &                     & DSDS-KMS            &           & 0.2370 $\pm$ 0.0006            &           & 0.2166 $\pm$ 0.0009            &       &         & DSDS-KMS    &       & 0.2301 $\pm$ 0.0010                  &       & 0.2102 $\pm$ 0.0025             \\
                                               & ResNet50-IN          & DSDS-KM            &           & 0.2384 $\pm$ 0.0006            &           & 0.2160 $\pm$ 0.0004            &       & Dino-B  & DSDS-KM     &       & 0.2261 $\pm$ 0.0017                  &       & 0.2100 $\pm$ 0.0022             \\
                                               &                     & DSDS-KMS            &           & 0.2390 $\pm$ 0.0002            &           & 0.2166 $\pm$ 0.0008            &       &         & DSDS-KMS    &       & 0.2280 $\pm$ 0.0014                  &       & 0.2111 $\pm$ 0.0044             \\
                                               & ResNeXt50-IN         & DSDS-KM            &           & 0.2389 $\pm$ 0.0002            &           & 0.2161 $\pm$ 0.0004            &       & Dino-L  & DSDS-KM     &       & 0.2283 $\pm$ 0.0014                  &       & 0.2104 $\pm$ 0.0011             \\
                                               &                     & DSDS-KMS            &           & 0.2389 $\pm$ 0.0004            &           & 0.2171 $\pm$ 0.0010            &       &         & DSDS-KMS    &       & 0.2290 $\pm$ 0.0009                  &       & 0.2126 $\pm$ 0.0023             \\
                                               & ViT-L-IN             & DSDS-KM            &           & 0.2382 $\pm$ 0.0003            &           & 0.2157 $\pm$ 0.0008            &       & Dino-G  & DSDS-KM     &       & \underline{0.2258 $\pm$ 0.0015}      &       & \underline{0.2099 $\pm$ 0.0018} \\
                                               &                     & DSDS-KMS            &           & 0.2381 $\pm$ 0.0002            &           & 0.2161 $\pm$ 0.0004            &       &         & DSDS-KMS    &       & 0.2260 $\pm$ 0.0015                  &       & 0.2108 $\pm$ 0.0018             \\
    \midrule
    \midrule
                                % &                  &                    &           &                                  &                                   &           &                                  &                                   \\
    \multirow{3}{*}{\rotatebox{90}{Static}}     &  & Passive            &           & \underline{0.2653}                &           & \underline{0.2383}                &       &       &  Passive     &       & \underline{0.2653}             &           & \underline{0.2383} \\
                                                &  & Rail               &           & 0.2716                            &           & 0.2453                            &       &       &  Rail        &       & 0.2716                            &           & 0.2453               \\
                                &                  & Sportive           &           & 0.2738                            &           & 0.2470                            &       &       &  Sportive    &       & 0.2738                            &           & 0.2470               \\
    \bottomrule 
    \end{tabular}
    \label{tab:imageNetDino}
    \end{table*}

\endgroup
To quantify if a pretraining on a task-specific pretrain dataset is necessary, we infer representations with models pretrained supervised on ImageNet1K and pretrained unsupervised on curated data from different sources.
Pretrained models on publicly available datasets alleviate the time and resource requirements for gathering a large-scale driving dataset.
Furthermore, for unsupervised learning, studies show beneficial characteristics of the learned representations, like the transferability to various target tasks \cite{ericsson2021well,zhao2020makes,sariyildiz2021concept,stuhr2022don, oquab2023dinov2} or the existence of more detailed information in the representation than supervised learning \cite{caron2021emerging,bordes2021high}.
Therefore, the representations of these models could have beneficial characteristics for situation-based clustering.
As seen in \autoref{tab:imageNetDino}, the overall performance of supervised and unsupervised pretraining is very similar.
This aligns with other studies \cite{ericsson2021well,zhao2020makes,sariyildiz2021concept, oquab2023dinov2} that show evidence that unsupervised pretraining can be competitive with supervised pretraining without requiring labeled data.
Additionally, no clear correlation is observed between the evaluated representation sizes and the resulting RMSE values.
However, compared to the task-specific pretraining results summarized in \autoref{tab:supervisedVal}, we observe a notable drop in performance.
This decrease in performance is similar for both the NN and DSC approaches, with DSC now only slightly outperforming the static driving styles.
According to robust ANOVAs, the mean differences of the errors remain statistically significant (\pValuePL{}) for both the visual feature encoders pretrained supervised on ImageNet1K \anovaPL{\robustANOVAName}{2.0}{6507}{685}{.001} and unsupervised on curated data \anovaPL{\robustANOVAName}{2.0}{6155}{1047}{.001}.
Although these pretrained feature encoders can lead to a more situation-specific clustering, as shown in \autoref{fig:clusterImages}, the observed drop in performance can be attributed to unwanted invariances or missing information required for driving behavior prediction in the representations.
One potential explanation for this can be drawn from the qualitative analysis of the situation cluster images, exemplarily shown in row four of \autoref{fig:clusterImages}.
Here it is indicated that the clusters trained on the representations obtained from the visual feature encoders pretrained on $\mathcal{D}_{P}$ are more sensitive to the road curvature.
In contrast, the other visual feature encoders focus more on the general visual appearance of the scene.
Overall, it is evident that all representations obtained from the different variants of feature encoders are able to form plausible situation clusters. 
However, there are possible shortcomings, such as unclear driving situations or over-specification,  as highlighted in the last row of \autoref{fig:clusterImages}.

\subsection*{Cluster Specificity}
\begin{figure}[t]
    \centering
    \small
    \begin{tabular}{cc}
        \includegraphics[trim={0cm 0 -0.75cm 0}, clip, width=.47\linewidth,valign=m]{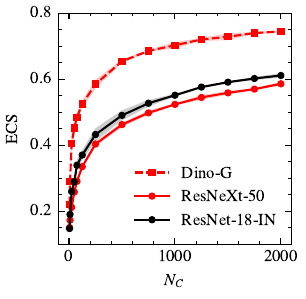}
        & 
        \includegraphics[trim={0cm 0 -0.75cm 0}, clip,width=.47\linewidth,valign=m]{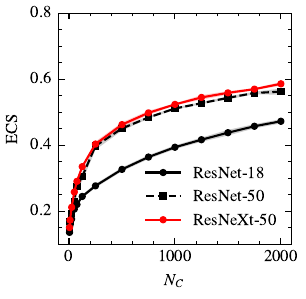}
        \\ 
        a) & b) \\
    \end{tabular}
    \caption{
        a) Comparison of the Entropy-based Cluster Specificity ($\mathrm{ECS}$) over the number of clusters $N_C$ of the best-performing models for pretraining variant.
        b) ECS curves for the models pretrained on our pretrain dataset $\mathcal{D}_{P,T}$.
    }
\label{fig:clusterSpeci}
\end{figure}

To quantitatively analyze the specificity of the found situation clusters, we utilize our proposed ECS metric for the clustered representations obtained from the different visual feature encoders. 
As seen in \autoref{fig:clusterSpeci} a), the visual feature encoders pretrained supervised on ImageNet1K and unsupervised on curated data achieve higher specificity compared to the visual feature encoders pretrained on our dataset $\mathcal{D}_{P,T}$.
Generally, we observe increasing specificity values for an increasing number of clusters $N_C$ and stable specificity results across multiple runs in our experiments.
The unsupervised Dinov2 models lead to the highest specificity, even for a lower number of clusters.
This high specificity is also visible in the cluster image samples of \autoref{fig:clusterImages}, where the high ECS scores underline the ability to differentiate driving situations in detail.
However, a higher specificity can lead to a decrease in generalization and does not generally correlate with a good performance on a target task like behavior prediction.
This can be seen in the higher RMSE values of the Dinov2 models and the models pretrained on ImageNet1K.
Therefore, archiving a high precision on the target task (generalization) while maintaining high specificity is beneficial for our method.
In our experiments, such a trend can be observed for the visual feature encoders trained on our pretrain dataset, as shown in \autoref{fig:clusterSpeci} b), where higher-performing models also exhibit higher specificity.

\subsection*{Iterative Driving Style Adaptation}

\begin{figure*}[]
    \centering
    \small
    \begin{tabular}{cccc}
        % ~ & \multicolumn{3}{c}{\textbf{Feature Encoder Pretrained on}} \\[0.2cm]
        ~ & \textbf{Feature Encoder Pretrained on} & \textbf{Feature Encoder Pretrained on} & \textbf{Feature Encoder Pretrained on} \\
        ~ & \textbf{$\mathcal{D}_{P}$ (ours)} & \textbf{ImageNet1K} & \textbf{Currated Data (Dinov2)} \\
        % 10 ITERATIONS
        \rotatebox[origin=c]{90}{\textbf{~ 10 \%}} &
        \includegraphics[trim={0cm 1.2cm -0.65cm -0.2cm}, clip, width=.25\linewidth,valign=m]{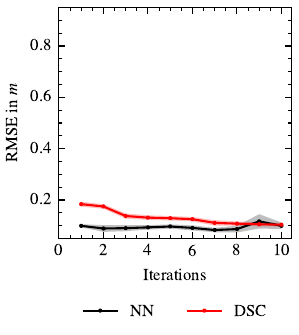} & 
        \includegraphics[trim={0cm 1.2cm -0.65cm -0.2cm}, clip,width=.25\linewidth,valign=m]{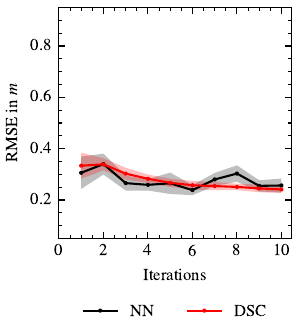} & 
        \includegraphics[trim={0cm 1.2cm -0.65cm -0.2cm}, clip,width=.25\linewidth,valign=m]{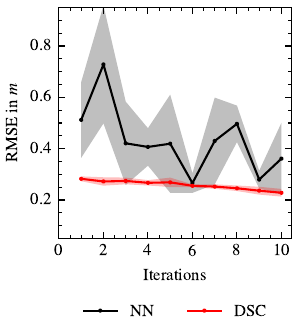} \\
        % 100 ITERATIONS
        \rotatebox[origin=c]{90}{\textbf{~ 1 \%}} &
        \includegraphics[trim={0cm 1.2cm -0.65cm -0.2cm}, clip, width=.25\linewidth,valign=m]{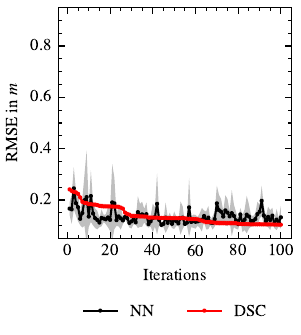} & 
        \includegraphics[trim={0cm 1.2cm -0.65cm -0.2cm}, clip,width=.25\linewidth,valign=m]{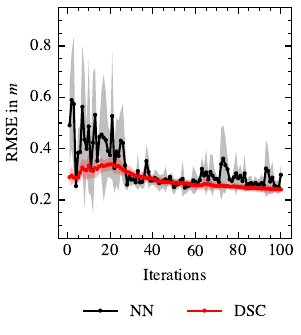} & 
        \includegraphics[trim={0cm 1.2cm -0.65cm -0.2cm}, clip,width=.25\linewidth,valign=m]{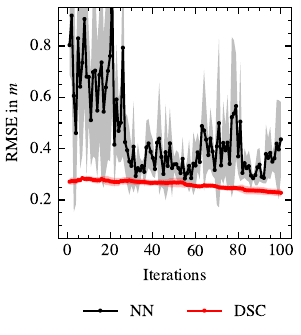} \\ 
        % 200 ITERATIONS
        \rotatebox[origin=c]{90}{\textbf{~~~ 0.5 \%}} &
        \includegraphics[trim={0cm 0.8cm -0.65cm -0.2cm}, clip, width=.25\linewidth,valign=m]{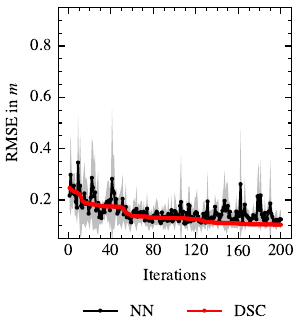} & 
        \includegraphics[trim={0cm 0.8cm -0.65cm -0.2cm}, clip,width=.25\linewidth,valign=m]{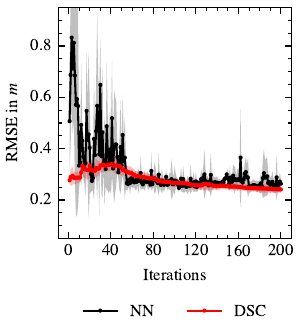} & 
        \includegraphics[trim={0cm 0.8cm -0.65cm -0.2cm}, clip,width=.25\linewidth,valign=m]{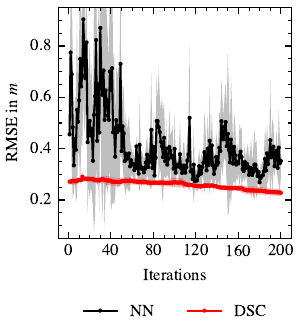} \\ 
        ~ & ~ & ~ & ~ \\
        ~ & \multicolumn{3}{c}{
            \includegraphics[trim={0cm 0cm -0.65cm 5cm}, clip,width=.34\linewidth,valign=m]{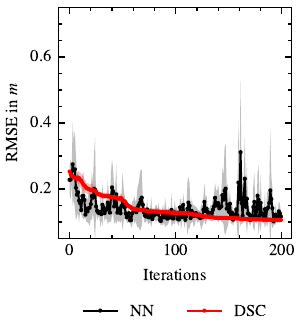}
        }
    \end{tabular}
    \caption{
        Training curves for the NN and DSC-based driving behavior prediction when trained iteratively on subsets of $\mathcal{D}_{V,T}$.
        The subset sizes are \SI{10}{\percent}, \SI{1}{\percent}, and \SI{0.5}{\percent} of the training data $\mathcal{D}_{V,T}$.
        The performance based on feature encoders pretrained on our pretrain dataset $\mathcal{D}_{P,T}$, ImageNet1K and in an unsupervised manner on curated data from different sources is shown.
    }
  \label{fig:trainIter}
  \end{figure*}

To evaluate the capability of our method to adapt to the driving style of a specific driver synchronously while gathering driving data, we split the dataset $\mathcal{D}_{V,T}$ into smaller subsets.
We maintain the temporal order of the driving data when splitting into these subsets to mirror a real-world recording.
For each training iteration, the models are trained on the respective subset until all driving data has been processed.
We experiment with subsets that contain \SI{10}{\percent}, \SI{1}{\percent}, and \SI{0.5}{\percent} of the training dataset $\mathcal{D}_{V,T}$.
For each iteration, we validate our models using the entire validation set $\mathcal{D}_{V,V}$ to show overall improvements during the iterative training.
The training curves, visualized in \autoref{fig:trainIter}, show that the DSC approach converges to the same RMSE as when trained on the entire dataset $\mathcal{D}_{V,T}$ at once.
This behavior is expected since the lookup table training eliminates catastrophic forgetting by design, as the calculation of the statistics leads to the identical lookup table entries when training on the dataset iteratively or when training on the entire dataset $\mathcal{D}_{V,T}$.
Furthermore, the lookup table approach is low in training time and memory consumption since only the number of assigned samples and their sum need to be saved for each situation cluster.  
In contrast, for the MLP-based driving behavior prediction, catastrophic forgetting can be observed.
After the initial gains achieved by using the learned model from the previous iteration as initialization for the current iteration, no further increase in performance is visible.
However, after seeing only a few training samples, the performance of the MLP increases significantly and is outperformed by the fully-trained lookup table only by a small margin.
The MLP's capability to learn from a small number of samples and the performance variations among different pretrained feature encoders implies that the information embedded into the feature encoder significantly impacts the performance of behavior prediction.

\section{Conclusion}
This work shows that a situation-aware prediction of human driving behavior based on camera images that capture the driving environment significantly surpasses the performance of several static driving styles.
Moreover, a driving style adaptation based on visual feature encoders and situation clusters pretrained on fleet data results in a precise driving behavior modeling of different drivers with an average RMSE of \SI{6.77}{\centi\metre}.
This shows that a setup with a visual feature encoder pretrained, e.g., by the manufacturer, and with decoupled driver-specific prediction heads, like MLP- and driving-situation-clustering-based models, is feasible.
Furthermore, we experiment with visual feature encoders pretrained on other datasets to evaluate the need for dedicated task-specific pretraining datasets.
The qualitative results show that the different visual feature encoders focus on different aspects of driving situations.
To analyze these aspects quantitatively, we introduce an entropy-based cluster specificity metric. Using this metric, we observe that visual feature encoders pretrained on other datasets exhibit higher specificity values.
It is important to note that cluster specificity does not necessarily correlate with performance, and overspecialization on unrelated aspects could negatively impact driving behavior prediction.
However, a positive trend between higher specificity and a lower RMSE value for driving behaviour modeling can be observed for the visual feature encoders pretrained on our dataset. 
From a manufacturer's point of view, higher specificity values could prove advantageous in constraining and controlling driving style adaptation for specific situations with greater detail.
Therefore, a two-branched version of our method with a branch for behavior prediction and a branch for situation masking could be realized with two different visual feature encoders.
For an application-oriented test we evaluate the model's capability to be trained synchronously while gathering driving data.
While the MLP-based behavior predictors achieve good performance initially, they suffer from catastrophic forgetting and are unable to learn from a continuous data stream.
In contrast, the driving situation-dependent statistics can iteratively learn from the new driving samples by design. 
Overall, we found that the underlying visual feature encoder significantly impacts the performance of the driving behavior prediction, indicating that relevant information for driving behavior prediction is contained within situation-dependent representations.

\subsection*{Limitations}
A potential limitation of our work is the usage of a single image for behavior prediction, which could be extended in future work into a sequence-based approach to incorporate the temporal information into the predictions.
Our proposed publicly available dataset is already suitable for temporal methods.
Furthermore, driving behavior predictors can be improved by utilizing more advanced models than MPLs or by improving the situation clustering and the statistical inference of the DSC approach.
One interesting direction would be to train separate prediction heads per situation cluster.
While our method can theoretically predict multiple driving behavior indicators, additional research needs to be conducted to explore other use cases, such as predicting longitudinal indicators suitable for Adaptive Cruise Control (ACC).
Additionally, it is important to highlight that the collection of data for autonomous driving is an ongoing effort, and datasets like ours do not encompass all possible real-world driving scenarios that are crucial to ensure safe and practical deployment.
Although the results show significant improvements compared to static driving styles, there is still a need for a more profound understanding of how sensitive human driving behavior is regarding variations in distances to the lane center.

% \noindent\textbf{Ethical and Responsible Use.}
\subsection*{Ethical and Responsible Use}
Overall, our work contributes to the ongoing research in the field of autonomous driving, which still deals with unresolved ethical and legal questions.
Our method intends to adapt behavior predictors to the driving style of different drivers live during driving.
While a live adaptation should be treated with caution, we mitigate possible risks by decoupling
driving behavior indicators from the actual control quantities.
This enables a driving style adaptation safeguarded by the low-level controller.
Furthermore, considering the limitations of our dataset, real-world tests should be conducted with care in a safe environment.
To publish the data concerning privacy policies, we utilized a state-of-the-art anonymization framework to blur human faces and vehicle license plates to mitigate privacy concerns.

\section*{Acknowledgments}
This work is being conducted as part of a research project of the Institute for Driver Assistance Systems and Connected Mobility (IFM) of the Allgäu Research Center at the University of Applied Sciences Kempten.

% {\appendix[Proof of the Zonklar Equations]
% Use $\backslash${\tt{appendix}} if you have a single appendix:
% Do not use $\backslash${\tt{section}} anymore after $\backslash${\tt{appendix}}, only $\backslash${\tt{section*}}.
% If you have multiple appendixes use $\backslash${\tt{appendices}} then use $\backslash${\tt{section}} to start each appendix.
% You must declare a $\backslash${\tt{section}} before using any $\backslash${\tt{subsection}} or using $\backslash${\tt{label}} ($\backslash${\tt{appendices}} by itself
%  starts a section numbered zero.)}

%{\appendices
%\section*{Proof of the First Zonklar Equation}
%Appendix one text goes here.
% You can choose not to have a title for an appendix if you want by leaving the argument blank
%\section*{Proof of the Second Zonklar Equation}
%Appendix two text goes here.}

%\section{References Section}
%You can use a bibliography generated by BibTeX as a .bbl file.
% BibTeX documentation can be easily obtained at:
% http://mirror.ctan.org/biblio/bibtex/contrib/doc/
% The IEEEtran BibTeX style support page is:
% http://www.michaelshell.org/tex/ieeetran/bibtex/
% 
% % argument is your BibTeX string definitions and bibliography database(s)
% \clearpage
\bibliographystyle{IEEEtran}
\bibliography{IEEEabrv,References.bib}

% Generated by IEEEtran.bst, version: 1.14 (2015/08/26)
\begin{thebibliography}{100}
\providecommand{\url}[1]{#1}
\csname url@samestyle\endcsname
\providecommand{\newblock}{\relax}
\providecommand{\bibinfo}[2]{#2}
\providecommand{\BIBentrySTDinterwordspacing}{\spaceskip=0pt\relax}
\providecommand{\BIBentryALTinterwordstretchfactor}{4}
\providecommand{\BIBentryALTinterwordspacing}{\spaceskip=\fontdimen2\font plus
\BIBentryALTinterwordstretchfactor\fontdimen3\font minus
  \fontdimen4\font\relax}
\providecommand{\BIBforeignlanguage}[2]{{%
\expandafter\ifx\csname l@#1\endcsname\relax
\typeout{** WARNING: IEEEtran.bst: No hyphenation pattern has been}%
\typeout{** loaded for the language `#1'. Using the pattern for}%
\typeout{** the default language instead.}%
\else
\language=\csname l@#1\endcsname
\fi
#2}}
\providecommand{\BIBdecl}{\relax}
\BIBdecl

\bibitem{bellem2018comfort}
H.~Bellem, B.~Thiel, M.~Schrauf, and J.~F. Krems, ``Comfort in automated
  driving: An analysis of preferences for different automated driving styles
  and their dependence on personality traits,'' \emph{Transportation research
  part F: traffic psychology and behaviour}, vol.~55, pp. 90--100, 2018.

\bibitem{ekman2019exploring}
F.~Ekman, M.~Johansson, L.-O. Blig{\aa}rd, M.~Karlsson, and H.~Str{\"o}mberg,
  ``Exploring automated vehicle driving styles as a source of trust
  information,'' \emph{Transportation research part F: traffic psychology and
  behaviour}, vol.~65, pp. 268--279, 2019.

\bibitem{strauch2019real}
C.~Strauch, K.~M{\"u}hl, K.~Patro, C.~Grabmaier, S.~Reithinger, M.~Baumann, and
  A.~Huckauf, ``Real autonomous driving from a passenger’s perspective: Two
  experimental investigations using gaze behaviour and trust ratings in field
  and simulator,'' \emph{Transportation research part F: traffic psychology and
  behaviour}, vol.~66, pp. 15--28, 2019.

\bibitem{carsten2019can}
O.~Carsten and M.~H. Martens, ``How can humans understand their automated cars?
  hmi principles, problems and solutions,'' \emph{Cognition, Technology \&
  Work}, vol.~21, no.~1, pp. 3--20, 2019.

\bibitem{ramm2014first}
S.~Ramm, J.~Giacomin, D.~Robertson, and A.~Malizia, ``A first approach to
  understanding and measuring naturalness in driver-car interaction,'' in
  \emph{Proceedings of the 6th International Conference on Automotive User
  Interfaces and Interactive Vehicular Applications}, 2014, pp. 1--10.

\bibitem{martinez2017driving}
C.~M. Martinez, M.~Heucke, F.-Y. Wang, B.~Gao, and D.~Cao, ``Driving style
  recognition for intelligent vehicle control and advanced driver assistance: A
  survey,'' \emph{IEEE Transactions on Intelligent Transportation Systems},
  vol.~19, no.~3, pp. 666--676, 2017.

\bibitem{sun2017research}
B.~Sun, W.~Deng, J.~Wu, Y.~Li, B.~Zhu, and L.~Wu, ``Research on the
  classification and identification of driver’s driving style,'' in
  \emph{2017 10th International Symposium on Computational Intelligence and
  Design (ISCID)}, vol.~1.\hskip 1em plus 0.5em minus 0.4em\relax IEEE, 2017,
  pp. 28--32.

\bibitem{bruck2021investigation}
Y.~Br{\"u}ck, D.~Niermann, A.~Trende, and A.~L{\"u}dtke, ``Investigation of
  personality traits and driving styles for individualization of autonomous
  vehicles,'' in \emph{Intelligent Human Systems Integration 2021: Proceedings
  of the 4th International Conference on Intelligent Human Systems Integration
  (IHSI 2021): Integrating People and Intelligent Systems, February 22-24,
  2021, Palermo, Italy}.\hskip 1em plus 0.5em minus 0.4em\relax Springer, 2021,
  pp. 78--83.

\bibitem{drewitz2020towards}
U.~Drewitz, K.~Ihme, C.~Bahnm{\"u}ller, T.~Fleischer, H.~La, A.-A. Pape,
  D.~Gr{\"a}fing, D.~Niermann, and A.~Trende, ``Towards user-focused vehicle
  automation: the architectural approach of the autoakzept project,'' in
  \emph{HCI in Mobility, Transport, and Automotive Systems. Automated Driving
  and In-Vehicle Experience Design: Second International Conference, MobiTAS
  2020, Held as Part of the 22nd HCI International Conference, HCII 2020,
  Copenhagen, Denmark, July 19--24, 2020, Proceedings, Part I 22}.\hskip 1em
  plus 0.5em minus 0.4em\relax Springer, 2020, pp. 15--30.

\bibitem{chen2019driving}
K.-T. Chen and H.-Y.~W. Chen, ``Driving style clustering using naturalistic
  driving data,'' \emph{Transportation research record}, vol. 2673, no.~6, pp.
  176--188, 2019.

\bibitem{pion2012fingerprint}
O.~Pion, R.~Henze, and F.~K{\"u}{\c{c}}{\"u}kay, ``Fingerprint des fahrers zur
  adaption von assistenzsystemen,'' \emph{INFORMATIK 2012}, 2012.

\bibitem{van2018relation}
H.~H. Van~Huysduynen, J.~Terken, and B.~Eggen, ``The relation between
  self-reported driving style and driving behaviour. a simulator study,''
  \emph{Transportation research part F: traffic psychology and behaviour},
  vol.~56, pp. 245--255, 2018.

\bibitem{sun2020intention}
B.~Sun, W.~Deng, J.~Wu, Y.~Li, and J.~Wang, ``An intention-aware and online
  driving style estimation based personalized autonomous driving strategy,''
  \emph{International journal of automotive technology}, vol.~21, pp.
  1431--1446, 2020.

\bibitem{gkartzonikas2019have}
C.~Gkartzonikas and K.~Gkritza, ``What have we learned? a review of stated
  preference and choice studies on autonomous vehicles,'' \emph{Transportation
  Research Part C: Emerging Technologies}, vol.~98, pp. 323--337, 2019.

\bibitem{buyukyildiz2017identification}
G.~B{\"u}y{\"u}kyildiz, O.~Pion, C.~Hildebrandt, M.~Sedlmayr, R.~Henze, and
  F.~K{\"u}{\c{c}}{\"u}kay, ``Identification of the driving style for the
  adaptation of assistance systems,'' \emph{International Journal of Vehicle
  Autonomous Systems}, vol.~13, no.~3, pp. 244--260, 2017.

\bibitem{inagaki2003adaptive}
T.~Inagaki \emph{et~al.}, ``Adaptive automation: Sharing and trading of
  control,'' \emph{Handbook of cognitive task design}, vol.~8, pp. 147--169,
  2003.

\bibitem{bar2011probabilistic}
T.~B{\"a}r, D.~Nienh{\"u}ser, R.~Kohlhaas, and J.~M. Z{\"o}llner,
  ``Probabilistic driving style determination by means of a situation based
  analysis of the vehicle data,'' in \emph{2011 14th International IEEE
  Conference on Intelligent Transportation Systems (ITSC)}.\hskip 1em plus
  0.5em minus 0.4em\relax IEEE, 2011, pp. 1698--1703.

\bibitem{chu2017curve}
D.~Chu, Z.~Deng, Y.~He, C.~Wu, C.~Sun, and Z.~Lu, ``Curve speed model for
  driver assistance based on driving style classification,'' \emph{IET
  Intelligent Transport Systems}, vol.~11, no.~8, pp. 501--510, 2017.

\bibitem{karlsson2021encoding}
J.~Karlsson, S.~van Waveren, C.~Pek, I.~Torre, I.~Leite, and J.~Tumova,
  ``Encoding human driving styles in motion planning for autonomous vehicles,''
  in \emph{2021 IEEE International Conference on Robotics and Automation
  (ICRA)}.\hskip 1em plus 0.5em minus 0.4em\relax IEEE, 2021, pp. 1050--1056.

\bibitem{phinnemore2021happy}
R.~Phinnemore, G.~Cimolino, P.~Sarkar, A.~Etemad, and T.~N. Graham, ``Happy
  driver: Investigating the effect of mood on preferred style of driving in
  self-driving cars,'' in \emph{Proceedings of the 9th International Conference
  on Human-Agent Interaction}, 2021, pp. 139--147.

\bibitem{ma2021drivers}
Z.~Ma and Y.~Zhang, ``Drivers trust, acceptance, and takeover behaviors in
  fully automated vehicles: Effects of automated driving styles and driver’s
  driving styles,'' \emph{Accident Analysis \& Prevention}, vol. 159, p.
  106238, 2021.

\bibitem{itkonen2020characterisation}
T.~H. Itkonen, E.~Lehtonen \emph{et~al.}, ``Characterisation of motorway
  driving style using naturalistic driving data,'' \emph{Transportation
  research part F: traffic psychology and behaviour}, vol.~69, pp. 72--79,
  2020.

\bibitem{chu2020self}
H.~Chu, L.~Guo, Y.~Yan, B.~Gao, and H.~Chen, ``Self-learning optimal cruise
  control based on individual car-following style,'' \emph{IEEE Transactions on
  Intelligent Transportation Systems}, vol.~22, no.~10, pp. 6622--6633, 2020.

\bibitem{chen2021driving}
D.~Chen, Z.~Chen, Y.~Zhang, X.~Qu, M.~Zhang, and C.~Wu, ``Driving style
  recognition under connected circumstance using a supervised hierarchical
  bayesian model,'' \emph{Journal of advanced transportation}, vol. 2021, pp.
  1--12, 2021.

\bibitem{elander1993behavioral}
J.~Elander, R.~West, and D.~French, ``Behavioral correlates of individual
  differences in road-traffic crash risk: An examination of methods and
  findings.'' \emph{Psychological bulletin}, vol. 113, no.~2, p. 279, 1993.

\bibitem{lajunen2011self}
T.~Lajunen and T.~{\"O}zkan, ``Self-report instruments and methods,'' in
  \emph{Handbook of traffic psychology}.\hskip 1em plus 0.5em minus 0.4em\relax
  Elsevier, 2011, pp. 43--59.

\bibitem{sagberg2015review}
F.~Sagberg, Selpi, G.~F. Bianchi~Piccinini, and J.~Engstr{\"o}m, ``A review of
  research on driving styles and road safety,'' \emph{Human factors}, vol.~57,
  no.~7, pp. 1248--1275, 2015.

\bibitem{kleisen2011relationship}
L.~Kleisen, \emph{The relationship between thinking and driving styles and
  their contribution to young driver road safety}.\hskip 1em plus 0.5em minus
  0.4em\relax University of Canberra Bruce, Australia, 2011.

\bibitem{tement2022assessment}
S.~Tement, B.~Musil, N.~Plohl, M.~Horvat, K.~Stojmenova, and J.~Sodnik,
  ``Assessment and profiling of driving style and skills,'' \emph{User
  Experience Design in the Era of Automated Driving}, pp. 151--176, 2022.

\bibitem{hasenjager2019survey}
M.~Hasenj{\"a}ger, M.~Heckmann, and H.~Wersing, ``A survey of personalization
  for advanced driver assistance systems,'' \emph{IEEE Transactions on
  Intelligent Vehicles}, vol.~5, no.~2, pp. 335--344, 2019.

\bibitem{festner2016einfluss}
M.~Festner, H.~Baumann, and D.~Schramm, ``Der einfluss fahrfremder
  t{\"a}tigkeiten und man{\"o}verl{\"a}ngsdynamik auf die komfort-und
  sicherheitswahrnehmung beim hochautomatisierten fahren,''
  \emph{VW-Gemeinschaftstagung Fahrerassistenz und automatisiertes Fahren,
  Wolfsburg}, 2016.

\bibitem{griesche2016should}
S.~Griesche, E.~Nicolay, D.~Assmann, M.~Dotzauer, and D.~K{\"a}thner, ``Should
  my car drive as i do? what kind of driving style do drivers prefer for the
  design of automated driving functions,'' in \emph{Braunschweiger Symposium},
  vol.~10, no.~11, 2016, pp. 185--204.

\bibitem{bolduc2019multimodel}
A.~P. Bolduc, L.~Guo, and Y.~Jia, ``Multimodel approach to personalized
  autonomous adaptive cruise control,'' \emph{IEEE Transactions on Intelligent
  Vehicles}, vol.~4, no.~2, pp. 321--330, 2019.

\bibitem{sun2020exploring}
X.~Sun, J.~Li, P.~Tang, S.~Zhou, X.~Peng, H.~N. Li, and Q.~Wang, ``Exploring
  personalised autonomous vehicles to influence user trust,'' \emph{Cognitive
  Computation}, vol.~12, pp. 1170--1186, 2020.

\bibitem{hartwich2015drive}
F.~Hartwich, M.~Beggiato, A.~Dettmann, and J.~F. Krems, ``Drive me comfortable:
  customized automated driving styles for younger and older drivers. 8,''
  \emph{VDI-Tagung “Der Fahrer im}, vol.~21, pp. 442--456, 2015.

\bibitem{rossner2022also}
P.~Rossner, M.~Friedrich, and A.~Bullinger-Hoffmann, \emph{I also care in
  manual driving - Influence of type, position and quantity of oncoming
  vehicles on manual driving behaviour in curves on rural roads}, 07 2022, pp.
  75 -- 85.

\bibitem{dettmann2021comfort}
A.~Dettmann, F.~Hartwich, P.~Ro{\ss}ner, M.~Beggiato, K.~Felbel, J.~Krems, and
  A.~C. Bullinger, ``Comfort or not? automated driving style and user
  characteristics causing human discomfort in automated driving,''
  \emph{International Journal of Human--Computer Interaction}, vol.~37, no.~4,
  pp. 331--339, 2021.

\bibitem{gao2020personalized}
B.~Gao, K.~Cai, T.~Qu, Y.~Hu, and H.~Chen, ``Personalized adaptive cruise
  control based on online driving style recognition technology and model
  predictive control,'' \emph{IEEE transactions on vehicular technology},
  vol.~69, no.~11, pp. 12\,482--12\,496, 2020.

\bibitem{ponomarev2019adaptation}
A.~Ponomarev and A.~Chernysheva, ``Adaptation and personalization in driver
  assistance systems,'' in \emph{2019 24th Conference of Open Innovations
  Association (FRUCT)}.\hskip 1em plus 0.5em minus 0.4em\relax IEEE, 2019, pp.
  335--344.

\bibitem{rosenfeld2012learning}
A.~Rosenfeld, Z.~Bareket, C.~Goldman, S.~Kraus, D.~LeBlanc, and O.~Tsimoni,
  ``Learning driver’s behavior to improve the acceptance of adaptive cruise
  contr,'' in \emph{Proceedings of the AAAI Conference on Artificial
  Intelligence}, vol.~26, no.~2, 2012, pp. 2317--2322.

\bibitem{rosenfeld2012towards}
A.~Rosenfeld, Z.~Bareket, C.~V. Goldman, S.~Kraus, D.~J. LeBlanc, and
  O.~Tsimhoni, ``Towards adapting cars to their drivers,'' \emph{AI Magazine},
  vol.~33, no.~4, pp. 46--46, 2012.

\bibitem{choi2021dsagands}
S.~Choi, N.~Kweon, C.~Yang, D.~Kim, H.~Shon, J.~Choi, and K.~Huh, ``Dsa-gan:
  Driving style attention generative adversarial network for vehicle trajectory
  prediction,'' \emph{2021 IEEE International Intelligent Transportation
  Systems Conference (ITSC)}, pp. 1515--1520, 2021.

\bibitem{lv2019drivingstylebasedco}
C.~Lv, X.~Hu, A.~Sangiovanni-Vincentelli, Y.~Li, C.~M. Martinez, and D.~Cao,
  ``Driving-style-based codesign optimization of an automated electric vehicle:
  A cyber-physical system approach,'' \emph{IEEE Transactions on Industrial
  Electronics}, vol.~66, pp. 2965--2975, 2019.

\bibitem{mohammadnazar2021classifyingtd}
A.~Mohammadnazar, R.~Arvin, and A.~Khattak, ``Classifying travelers' driving
  style using basic safety messages generated by connected vehicles:
  Application of unsupervised machine learning,'' \emph{Transportation Research
  Part C-emerging Technologies}, vol. 122, p. 102917, 2021.

\bibitem{khodairy2021drivingbc}
M.~A. Khodairy and G.~Abosamra, ``Driving behavior classification based on
  oversampled signals of smartphone embedded sensors using an optimized
  stacked-lstm neural networks,'' \emph{IEEE Access}, vol.~9, pp. 4957--4972,
  2021.

\bibitem{kovaceva2020identificationoa}
J.~Kovaceva, I.~Isaksson-Hellman, and N.~Murgovski, ``Identification of
  aggressive driving from naturalistic data in car-following situations.''
  \emph{Journal of safety research}, vol.~73, pp. 225--234, {2020}.

\bibitem{kim2021drivingsc}
D.~Kim, H.~Shon, N.~Kweon, S.~Choi, C.~Yang, and K.~Huh, ``Driving style-based
  conditional variational autoencoder for prediction of ego vehicle
  trajectory,'' \emph{IEEE Access}, vol.~PP, pp. 1--1, 2021.

\bibitem{dong2016characterizing}
W.~Dong, J.~Li, R.~Yao, C.~Li, T.~Yuan, and L.~Wang, ``Characterizing driving
  styles with deep learning,'' \emph{arXiv preprint arXiv:1607.03611}, 2016.

\bibitem{shouno2018deep}
O.~Shouno, ``Deep unsupervised learning of a topological map of vehicle
  maneuvers for characterizing driving styles,'' in \emph{2018 21st
  International Conference on Intelligent Transportation Systems (ITSC)}.\hskip
  1em plus 0.5em minus 0.4em\relax IEEE, 2018, pp. 2917--2922.

\bibitem{han2019statistical}
W.~Han, W.~Wang, X.~Li, and J.~Xi, ``Statistical-based approach for driving
  style recognition using bayesian probability with kernel density
  estimation,'' \emph{IET Intelligent Transport Systems}, vol.~13, no.~1, pp.
  22--30, 2019.

\bibitem{ghasemzadeh2018utilizing}
A.~Ghasemzadeh and M.~M. Ahmed, ``Utilizing naturalistic driving data for
  in-depth analysis of driver lane-keeping behavior in rain: Non-parametric
  mars and parametric logistic regression modeling approaches,''
  \emph{Transportation research part C: emerging technologies}, vol.~90, pp.
  379--392, 2018.

\bibitem{constantinescu2010driving}
Z.~Constantinescu, C.~Marinoiu, and M.~Vladoiu, ``Driving style analysis using
  data mining techniques,'' \emph{International Journal of Computers
  Communications \& Control}, vol.~5, no.~5, pp. 654--663, 2010.

\bibitem{chen2019graphical}
C.~Chen, X.~Zhao, Y.~Zhang, J.~Rong, and X.~Liu, ``A graphical modeling method
  for individual driving behavior and its application in driving safety
  analysis using gps data,'' \emph{Transportation research part F: traffic
  psychology and behaviour}, vol.~63, pp. 118--134, 2019.

\bibitem{hamdar2016weather}
S.~H. Hamdar, L.~Qin, and A.~Talebpour, ``Weather and road geometry impact on
  longitudinal driving behavior: Exploratory analysis using an empirically
  supported acceleration modeling framework,'' \emph{Transportation research
  part C: emerging technologies}, vol.~67, pp. 193--213, 2016.

\bibitem{robert2009individual}
L.~P. Robert, A.~R. Denis, and Y.-T.~C. Hung, ``Individual swift trust and
  knowledge-based trust in face-to-face and virtual team members,''
  \emph{Journal of management information systems}, vol.~26, no.~2, pp.
  241--279, 2009.

\bibitem{petersen2019situational}
L.~Petersen, L.~Robert, X.~J. Yang, and D.~M. Tilbury, ``Situational awareness,
  drivers trust in automated driving systems and secondary task performance,''
  \emph{arXiv preprint arXiv:1903.05251}, 2019.

\bibitem{ahmed2018impacts}
M.~M. Ahmed and A.~Ghasemzadeh, ``The impacts of heavy rain on speed and
  headway behaviors: An investigation using the shrp2 naturalistic driving
  study data,'' \emph{Transportation research part C: emerging technologies},
  vol.~91, pp. 371--384, 2018.

\bibitem{kilpelainen2007effects}
M.~Kilpel{\"a}inen and H.~Summala, ``Effects of weather and weather forecasts
  on driver behaviour,'' \emph{Transportation research part F: traffic
  psychology and behaviour}, vol.~10, no.~4, pp. 288--299, 2007.

\bibitem{rahman2012analysis}
A.~Rahman and N.~E. Lownes, ``Analysis of rainfall impacts on platooned vehicle
  spacing and speed,'' \emph{Transportation research part F: traffic psychology
  and behaviour}, vol.~15, no.~4, pp. 395--403, 2012.

\bibitem{faria2020assessing}
M.~V. Faria, P.~C. Baptista, T.~L. Farias, and J.~M. Pereira, ``Assessing the
  impacts of driving environment on driving behavior patterns,''
  \emph{Transportation}, vol.~47, no.~3, pp. 1311--1337, 2020.

\bibitem{hamada2016modeling}
R.~Hamada, T.~Kubo, K.~Ikeda, Z.~Zhang, T.~Shibata, T.~Bando, K.~Hitomi, and
  M.~Egawa, ``Modeling and prediction of driving behaviors using a
  nonparametric bayesian method with ar models,'' \emph{IEEE Transactions on
  Intelligent Vehicles}, vol.~1, no.~2, pp. 131--138, 2016.

\bibitem{rossner2020care}
P.~Rossner and A.~C. Bullinger, ``I care who and where you are--influence of
  type, position and quantity of oncoming vehicles on perceived safety during
  automated driving on rural roads,'' in \emph{HCI in Mobility, Transport, and
  Automotive Systems. Driving Behavior, Urban and Smart Mobility: Second
  International Conference, MobiTAS 2020, Held as Part of the 22nd HCI
  International Conference, HCII 2020, Copenhagen, Denmark, July 19--24, 2020,
  Proceedings, Part II 22}.\hskip 1em plus 0.5em minus 0.4em\relax Springer,
  2020, pp. 61--71.

\bibitem{bellem2017can}
H.~Bellem, M.~Kl{\"u}ver, M.~Schrauf, H.-P. Sch{\"o}ner, H.~Hecht, and J.~F.
  Krems, ``Can we study autonomous driving comfort in moving-base driving
  simulators? a validation study,'' \emph{Human factors}, vol.~59, no.~3, pp.
  442--456, 2017.

\bibitem{lex2017objektive}
C.~Lex, M.~Schabauer, M.~Semmer, J.~Holzinger, T.~Schl{\"o}micher, Z.~F.
  Magosi, A.~Eichberger, and I.~V. Koglbauer, ``Objektive erfassung und
  subjektive bewertung menschlicher trajektorienwahl in einer naturalistic
  driving study,'' in \emph{9. VDI-Fachtagung" Der Fahrer im 21. Jahrhundert":
  Der Mensch im Fokus technischer Innovationen}.\hskip 1em plus 0.5em minus
  0.4em\relax Springer-VDI-Verlag GmbH \& Co. KG, 2017, pp. 177--192.

\bibitem{schlag2015auswirkungen}
B.~Schlag, J.~Voigt, C.~Lippold, and K.~Enzfelder, ``Auswirkungen von
  querschnittsgestaltung und l{\"a}ngsgerichteten markierungen auf das
  fahrverhalten auf landstra{\ss}en,'' 2015.

\bibitem{rosey2009impact}
F.~Rosey, J.-M. Auberlet, O.~Moisan, and G.~Dupr{\'e}, ``Impact of narrower
  lane width: Comparison between fixed-base simulator and real data,''
  \emph{Transportation research record}, vol. 2138, no.~1, pp. 112--119, 2009.

\bibitem{triggs1997effect}
T.~J. Triggs, ``The effect of approaching vehicles on the lateral position of
  cars travelling on a two-lane rural road,'' \emph{Australian Psychologist},
  vol.~32, no.~3, pp. 159--163, 1997.

\bibitem{hamdar2016weatherar}
S.~H. Hamdar, L.~Qin, and A.~Talebpour, ``Weather and road geometry impact on
  longitudinal driving behavior: Exploratory analysis using an empirically
  supported acceleration modeling framework,'' \emph{Transportation Research
  Part C-emerging Technologies}, vol.~67, pp. 193--213, 2016.

\bibitem{ahmed2018theio}
M.~M. Ahmed and A.~Ghasemzadeh, ``The impacts of heavy rain on speed and
  headway behaviors: An investigation using the shrp2 naturalistic driving
  study data,'' \emph{Transportation Research Part C: Emerging Technologies},
  2018.

\bibitem{ghasemzadeh2018utilizingnd}
A.~Ghasemzadeh and M.~M. Ahmed, ``Utilizing naturalistic driving data for
  in-depth analysis of driver lane-keeping behavior in rain: Non-parametric
  mars and parametric logistic regression modeling approaches,''
  \emph{Transportation Research Part C-emerging Technologies}, vol.~90, pp.
  379--392, 2018.

\bibitem{cordero2020recognitionot}
J.~Cordero, J.~Aguilar, K.~Aguilar, D.~Chávez, and E.~Puerto, ``Recognition of
  the driving style in vehicle drivers,'' \emph{Sensors (Basel, Switzerland)},
  vol.~20, 2020.

\bibitem{rath2019personalisedlk}
J.~Rath, C.~Senouth, and J.~Popieul, ``Personalised lane keeping assist
  strategy: adaptation to driving style,'' \emph{IET Control Theory \&
  Applications}, 2019.

\bibitem{shahverdy2021driverbd}
M.~Shahverdy, M.~Fathy, R.~Berangi, and M.~Sabokrou, ``Driver behaviour
  detection using 1d convolutional neural networks,'' \emph{Electronics
  Letters}, 2021.

\bibitem{liu2021exploitingmd}
Z.~Liu, J.~Zheng, Z.~Gong, H.~Zhang, and K.~Wu, ``Exploiting multi-source data
  for adversarial driving style representation learning,'' pp. 491--508, 2021.

\bibitem{ghasemzadeh2017driversla}
A.~Ghasemzadeh and M.~M. Ahmed, ``Drivers’ lane-keeping ability in heavy
  rain: Preliminary investigation using shrp 2 naturalistic driving study
  data,'' \emph{Transportation Research Record}, vol. 2663, pp. 108 -- 99,
  2017.

\bibitem{oezguel2018afu}
O.~Özgül, M.~Çakir, M.~Tan, M.~Amasyali, and H.~T. Hayvaci, ``A fully
  unsupervised framework for scoring driving style,'' \emph{2018 International
  Conference on Intelligent Systems (IS)}, pp. 228--234, 2018.

\bibitem{chen2021drivingsr}
D.~Chen, Z.~Chen, Y.~Zhang, X.~Qu, M.~Zhang, and C.~Wu, ``Driving style
  recognition under connected circumstance using a supervised hierarchical
  bayesian model,'' \emph{Journal of Advanced Transportation}, 2021.

\bibitem{karlsson2021encodinghd}
J.~Karlsson, S.~V. Waveren, C.~Pek, I.~Torre, I.~Leite, and J.~Tumova,
  ``Encoding human driving styles in motion planning for autonomous vehicles,''
  \emph{2021 IEEE International Conference on Robotics and Automation (ICRA)},
  pp. 1050--1056, {2021}.

\bibitem{schrum2023mavericad}
M.~L. Schrum, E.~S. Sumner, M.~Gombolay, and A.~Best, ``Maveric: A data-driven
  approach to personalized autonomous driving,'' \emph{ArXiv}, vol.
  abs/2301.08595, 2023.

\bibitem{zheng2022realtimeds}
X.~Zheng, P.~Yang, D.~Duan, X.~Cheng, and L.~Yang, ``Real-time driving style
  classification based on short-term observations,'' \emph{IET Commun.},
  vol.~16, pp. 1393--1402, {2022}.

\bibitem{hajiseyedjavadi2021effectoe}
F.~Hajiseyedjavadi, E.~Boer, R.~Romano, E.~Paschalidis, C.~Wei, A.~Solernou,
  D.~Forster, and N.~Merat, ``Effect of environmental factors and individual
  differences on subjective evaluation of human-like and conventional automated
  vehicle controllers,'' \emph{SSRN Electronic Journal}, 2021.

\bibitem{gao2020personalizedac}
B.~Gao, K.~Cai, T.~Qu, Y.~Hu, and H.~Chen, ``Personalized adaptive cruise
  control based on online driving style recognition technology and model
  predictive control,'' \emph{IEEE Transactions on Vehicular Technology},
  vol.~69, pp. 12\,482--12\,496, 2020.

\bibitem{moukafih2019aggressivedd}
Y.~Moukafih, H.~Hafidi, and M.~Ghogho, ``Aggressive driving detection using
  deep learning-based time series classification,'' \emph{2019 IEEE
  International Symposium on INnovations in Intelligent SysTems and
  Applications (INISTA)}, pp. 1--5, 2019.

\bibitem{itkonen2020characterisationom}
T.~H. Itkonen, E.~Lehtonen, and Selpi, ``Characterisation of motorway driving
  style using naturalistic driving data,'' \emph{Transportation Research Part
  F: Traffic Psychology and Behaviour}, 2020.

\bibitem{xing2020personalizedvt}
Y.~Xing, C.~Lv, and D.~Cao, ``Personalized vehicle trajectory prediction based
  on joint time-series modeling for connected vehicles,'' \emph{IEEE
  Transactions on Vehicular Technology}, vol.~69, pp. 1341--1352, 2020.

\bibitem{shahverdy2020driverbd}
M.~Shahverdy, M.~Fathy, R.~Berangi, and M.~Sabokrou, ``Driver behavior
  detection and classification using deep convolutional neural networks,''
  \emph{Expert Syst. Appl.}, vol. 149, p. 113240, 2020.

\bibitem{zheng2023towardsdp}
L.~Zheng, J.~Poveda, J.~Mullen, S.~Revankar, and M.-C. Lin, ``Towards driving
  policies with personality: Modeling behavior and style in risky scenarios via
  data collection in virtual reality,'' \emph{ArXiv}, vol. abs/2303.04901,
  2023.

\bibitem{li2021combinedtp}
H.~Li, C.~Wu, D.~Chu, L.~Lu, and K.~Cheng, ``Combined trajectory planning and
  tracking for autonomous vehicle considering driving styles,'' \emph{IEEE
  Access}, vol.~9, pp. 9453--9463, 2021.

\bibitem{wang2020analysisot}
Z.~Wang, M.~Guan, J.~Lan, B.~Yang, T.~Kaizuka, J.~Taki, and K.~Nakano,
  ``Analysis of truck driver behavior to design different lane change styles in
  automated driving,'' \emph{ArXiv}, vol. abs/2012.15164, 2020.

\bibitem{yadav2021investigatingte}
A.~Yadav and N.~Velaga, ``Investigating the effects of driving environment and
  driver characteristics on drivers’ compliance with speed limits,''
  \emph{Traffic Injury Prevention}, vol.~22, pp. 201 -- 206, {2021}.

\bibitem{natarajan2022towardad}
M.~Natarajan, K.~Akash, and T.~Misu, ``Toward adaptive driving styles for
  automated driving with users' trust and preferences,'' \emph{2022 17th
  ACM/IEEE International Conference on Human-Robot Interaction (HRI)}, pp.
  940--944, 2022.

\bibitem{peralta2022amf}
R.~Peralta, I.~Becerra, U.~Ruiz, and R.~Murrieta-Cid, ``A methodology for
  generating driving styles for autonomous cars,'' \emph{Journal of Intelligent
  Transportation Systems}, vol.~28, pp. 120 -- 140, 2022.

\bibitem{ramezani-khansari2021comparingte}
E.~Ramezani-Khansari, M.~Tabibi, F.~M. Nejad, and M.~Mesbah, ``Comparing the
  effect of age, gender, and desired speed on car-following behavior by using
  driving simulator,'' \emph{Journal of Advanced Transportation}, 2021.

\bibitem{tement2022assessmentap}
S.~Tement, B.~Musil, N.~Plohl, M.~Horvat, K.~Stojmenova, and J.~Sodnik,
  ``Assessment and profiling of driving style and skills,'' \emph{Studies in
  Computational Intelligence}, {2022}.

\bibitem{he2022anid}
Y.~He, S.~Yang, X.~Zhou, and X.-Y. Lu, ``An individual driving behavior
  portrait approach for professional driver of hdvs with naturalistic driving
  data,'' \emph{Computational Intelligence and Neuroscience}, vol. 2022,
  {2022}.

\bibitem{magana2018amf}
V.~C. Magana, X.~G. Pañeda, A.~G. Tuero, L.~Pozueco, R.~García, D.~Melendi,
  and A.~Rionda, ``A method for making a fair evaluation of driving styles in
  different scenarios with recommendations for their improvement,'' \emph{IEEE
  Intelligent Transportation Systems Magazine}, vol.~13, pp. 136--148, 2018.

\bibitem{jardin2020rulebasedds}
P.~Jardin, I.~Moisidis, S.~H.~S. Zetina, and S.~Rinderknecht, ``Rule-based
  driving style classification using acceleration data profiles,'' \emph{2020
  IEEE 23rd International Conference on Intelligent Transportation Systems
  (ITSC)}, pp. 1--6, 2020.

\bibitem{rossner2020icw}
P.~Rossner and A.~C. Bullinger-Hoffmann, ``I care who and where you are -
  influence of type, position and quantity of oncoming vehicles on perceived
  safety during automated driving on rural roads,'' pp. 61--71, {2020}.

\bibitem{rath2019alk}
J.~Rath, C.~Sentouh, and J.~Popieul, ``A lane keeping assist design: Adaptation
  to driving style based on aggressiveness,'' \emph{2019 American Control
  Conference (ACC)}, pp. 5316--5321, 2019.

\bibitem{cartes2019effectod}
P.~Cartes, T.~Echaveguren, and P.~Álvarez, ``Effect of driving style on
  operating speed in crest vertical curves of two-lane highways,''
  \emph{Proceedings of the Institution of Civil Engineers - Transport}, 2019.

\bibitem{brück2021investigationop}
Y.~Brück, D.~Niermann, A.~Trende, and A.~Lüdtke, ``Investigation of
  personality traits and driving styles for individualization of autonomous
  vehicles,'' 2021, pp. 78--83.

\bibitem{feraud2020asm}
I.~S. Feraud and J.~Naranjo, ``A systematic methodology to evaluate prediction
  models for driving style classification,'' \emph{Sensors (Basel,
  Switzerland)}, vol.~20, 2020.

\bibitem{savelonas2020classificationod}
M.~Savelonas, S.~Karkanis, and E.~Spyrou, ``Classification of driving behaviour
  using short-term and long-term summaries of sensor data,'' \emph{2020 5th
  South-East Europe Design Automation, Computer Engineering, Computer Networks
  and Social Media Conference (SEEDA-CECNSM)}, pp. 1--4, 2020.

\bibitem{xue2019rapidds}
Q.~Xue, K.~Wang, J.~Lu, and Y.~Liu, ``Rapid driving style recognition in
  car-following using machine learning and vehicle trajectory data,''
  \emph{Journal of Advanced Transportation}, 2019.

\bibitem{liu2019researchoc}
Y.~Liu, J.~Wang, P.~Zhao, D.~Qin, and Z.~Chen, ``Research on classification and
  recognition of driving styles based on feature engineering,'' \emph{IEEE
  Access}, vol.~7, pp. 89\,245--89\,255, 2019.

\bibitem{bejani2018aca}
M.~M. Bejani and M.~Ghatee, ``A context aware system for driving style
  evaluation by an ensemble learning on smartphone sensors data,''
  \emph{Transportation Research Part C-emerging Technologies}, vol.~89, pp.
  303--320, 2018.

\bibitem{savelonas2020hybridtr}
M.~Savelonas, D.~Mantzekis, N.~Labiris, A.~Tsakiri, S.~Karkanis, and E.~Spyrou,
  ``Hybrid time-series representation for the classification of driving
  behaviour,'' \emph{2020 15th International Workshop on Semantic and Social
  Media Adaptation and Personalization (SMA}, pp. 1--6, 2020.

\bibitem{chen2021semitraj2graphif}
C.~Chen, Q.~Liu, X.~Wang, C.~Liao, and D.~Zhang, ``semi-traj2graph identifying
  fine-grained driving style with gps trajectory data via multi-task
  learning,'' \emph{IEEE Transactions on Big Data}, vol.~8, pp. 1550--1565,
  {2021}.

\bibitem{jaafer2020dataao}
A.~Jaafer, G.~Nilsson, and G.~Como, ``Data augmentation of imu signals and
  evaluation via a semi-supervised classification of driving behavior,''
  \emph{2020 IEEE 23rd International Conference on Intelligent Transportation
  Systems (ITSC)}, pp. 1--6, {2020}.

\bibitem{chen2019driverib}
J.~Chen, Z.~Wu, and J.~Zhang, ``Driver identification based on hidden feature
  extraction by using adaptive nonnegativity-constrained autoencoder,''
  \emph{Appl. Soft Comput.}, vol.~74, pp. 1--9, {2019}.

\bibitem{moosavi2021drivingsr}
S.~Moosavi, P.~Mahajan, S.~Parthasarathy, C.~Saunders-Chukwu, and R.~Ramnath,
  ``Driving style representation in convolutional recurrent neural network
  model of driver identification,'' \emph{ArXiv}, vol. abs/2102.05843, 2021.

\bibitem{bejani2020convolutionalnn}
M.~M. Bejani and M.~Ghatee, ``Convolutional neural network with adaptive
  regularization to classify driving styles on smartphones,'' \emph{IEEE
  Transactions on Intelligent Transportation Systems}, vol.~21, pp. 543--552,
  2020.

\bibitem{dong2016characterizingds}
W.~Dong, J.~Li, R.~Yao, C.~Li, T.~Yuan, and L.~Wang, ``Characterizing driving
  styles with deep learning,'' \emph{ArXiv}, vol. abs/1607.03611, 2016.

\bibitem{li2019drivingsc}
G.~Li, F.~Zhu, X.~Qu, B.~Cheng, S.~Li, and P.~Green, ``Driving style
  classification based on driving operational pictures,'' \emph{IEEE Access},
  vol.~7, pp. 90\,180--90\,189, 2019.

\bibitem{chen2019drivingsc}
K.-T. Chen and H.~Chen, ``Driving style clustering using naturalistic driving
  data,'' \emph{Transportation Research Record}, vol. 2673, pp. 176 -- 188,
  2019.

\bibitem{sun2020ania}
B.~Sun, W.~Deng, J.~Wu, Y.~Li, and J.~Wang, ``An intention-aware and online
  driving style estimation based personalized autonomous driving strategy,''
  \emph{International Journal of Automotive Technology}, vol.~21, pp. 1431 --
  1446, {2020}.

\bibitem{zhu2020clusteringds}
R.~Zhu and M.~Wüthrich, ``Clustering driving styles via image processing,''
  \emph{Annals of Actuarial Science}, vol.~15, pp. 276 -- 290, 2020.

\bibitem{shouno2018deepul}
O.~Shouno, ``Deep unsupervised learning of a topological map of vehicle
  maneuvers for characterizing driving styles,'' \emph{2018 21st International
  Conference on Intelligent Transportation Systems (ITSC)}, pp. 2917--2922,
  2018.

\bibitem{haselberger2023self}
J.~Haselberger, ``Self-perception versus objective driving behavior: Subject
  study of lateral vehicle guidance,'' \emph{Transportation Research Part F:
  Traffic Psychology and Behaviour}, 2024.

\bibitem{haselberger2022jupiter}
J.~Haselberger, M.~Pelzer, B.~Schick, and S.~M{\"u}ller, ``Jupiter--ros based
  vehicle platform for autonomous driving research,'' in \emph{2022 IEEE
  International Symposium on Robotic and Sensors Environments (ROSE)}.\hskip
  1em plus 0.5em minus 0.4em\relax IEEE, 2022, pp. 1--8.

\bibitem{raina2023egoblur}
N.~Raina, G.~Somasundaram, K.~Zheng, S.~Saarinen, J.~Messiner, M.~Schwesinger,
  L.~Pesqueira, I.~Prasad, E.~Miller, P.~Gupta \emph{et~al.}, ``Egoblur:
  Responsible innovation in aria,'' \emph{arXiv preprint arXiv:2308.13093},
  2023.

\bibitem{barendswaard2019acm}
S.~Barendswaard, D.~Pool, E.~Boer, and D.~Abbink, ``A classification method for
  driver trajectories during curve-negotiation,'' \emph{2019 IEEE International
  Conference on Systems, Man and Cybernetics (SMC)}, pp. 3729--3734, 2019.

\bibitem{haselberger2023exploring}
J.~Haselberger, ``Exploring the influence of driving context on lateral driving
  style preferences: A simulator-based study,'' \emph{Transportation Research
  Part F: Traffic Psychology and Behaviour}, 2024.

\bibitem{hofer2020attribute}
M.~H{\"o}fer, F.~Fuhr, B.~Schick, and P.~E. Pfeffer, ``Attribute-based
  development of driver assistance systems,'' in \emph{10th International
  Munich Chassis Symposium 2019: chassis. tech plus}.\hskip 1em plus 0.5em
  minus 0.4em\relax Springer, 2020, pp. 293--306.

\bibitem{russakovsky2015imagenet}
O.~Russakovsky, J.~Deng, H.~Su, J.~Krause, S.~Satheesh, S.~Ma, Z.~Huang,
  A.~Karpathy, A.~Khosla, M.~Bernstein \emph{et~al.}, ``Imagenet large scale
  visual recognition challenge,'' \emph{International journal of computer
  vision}, vol. 115, pp. 211--252, 2015.

\bibitem{woo2018dynamic}
H.~Woo, Y.~Ji, Y.~Tamura, Y.~Kuroda, T.~Sugano, Y.~Yamamoto, A.~Yamashita, and
  H.~Asama, ``Dynamic state estimation of driving style based on driving risk
  feature,'' \emph{International Journal of Automotive Engineering}, vol.~9,
  no.~1, pp. 31--38, 2018.

\bibitem{brambilla2017comparison}
M.~Brambilla, P.~Mascetti, and A.~Mauri, ``Comparison of different driving
  style analysis approaches based on trip segmentation over gps information,''
  in \emph{2017 IEEE International Conference on Big Data (Big Data)}.\hskip
  1em plus 0.5em minus 0.4em\relax IEEE, 2017, pp. 3784--3791.

\bibitem{lin2014overview}
N.~Lin, C.~Zong, M.~Tomizuka, P.~Song, Z.~Zhang, and G.~Li, ``An overview on
  study of identification of driver behavior characteristics for automotive
  control,'' \emph{Mathematical Problems in Engineering}, vol. 2014, 2014.

\bibitem{kim2021driving}
D.~Kim, H.~Shon, N.~Kweon, S.~Choi, C.~Yang, and K.~Huh, ``Driving style-based
  conditional variational autoencoder for prediction of ego vehicle
  trajectory,'' \emph{IEEE Access}, vol.~9, pp. 169\,348--169\,356, 2021.

\bibitem{bae2020selfdrivingla}
I.~Bae, J.~Moon, J.~Jhung, H.~Suk, T.~Kim, H.~Park, J.~Cha, J.~Kim, D.~Kim, and
  S.~Kim, ``Self-driving like a human driver instead of a robocar: Personalized
  comfortable driving experience for autonomous vehicles,'' \emph{ArXiv}, vol.
  abs/2001.03908, 2020.

\bibitem{jamovi2023jamovi}
\BIBentryALTinterwordspacing
T.~jamovi project. (2023) jamovi (version 2.3). [Online]. Available:
  \url{https://www.jamovi.org}
\BIBentrySTDinterwordspacing

\bibitem{wilcox1967indices}
A.~R. Wilcox, ``Indices of qualitative variation.'' Oak Ridge National Lab.,
  Tenn., Tech. Rep., 1967.

\bibitem{he2016deep}
K.~He, X.~Zhang, S.~Ren, and J.~Sun, ``Deep residual learning for image
  recognition,'' in \emph{Proceedings of the IEEE conference on computer vision
  and pattern recognition}, 2016, pp. 770--778.

\bibitem{xie2017aggregated}
S.~Xie, R.~Girshick, P.~Doll{\'a}r, Z.~Tu, and K.~He, ``Aggregated residual
  transformations for deep neural networks,'' in \emph{Proceedings of the IEEE
  conference on computer vision and pattern recognition}, 2017, pp. 1492--1500.

\bibitem{dosovitskiy2020image}
A.~Dosovitskiy, L.~Beyer, A.~Kolesnikov, D.~Weissenborn, X.~Zhai,
  T.~Unterthiner, M.~Dehghani, M.~Minderer, G.~Heigold, S.~Gelly \emph{et~al.},
  ``An image is worth 16x16 words: Transformers for image recognition at
  scale,'' \emph{arXiv preprint arXiv:2010.11929}, 2020.

\bibitem{darcet2023vision}
T.~Darcet, M.~Oquab, J.~Mairal, and P.~Bojanowski, ``Vision transformers need
  registers,'' \emph{arXiv preprint arXiv:2309.16588}, 2023.

\bibitem{macqueen1967classification}
J.~MacQueen, ``Classification and analysis of multivariate observations,'' in
  \emph{Proceedings of the 5th Berkeley Symposium on Mathematical Statistics
  and Probability}, 1967, pp. 281--297.

\bibitem{johnson2019billion}
J.~Johnson, M.~Douze, and H.~J{\'e}gou, ``Billion-scale similarity search with
  {GPUs},'' \emph{IEEE Transactions on Big Data}, vol.~7, no.~3, pp. 535--547,
  2019.

\bibitem{hendrycks2019augmix}
D.~Hendrycks, N.~Mu, E.~D. Cubuk, B.~Zoph, J.~Gilmer, and B.~Lakshminarayanan,
  ``Augmix: A simple data processing method to improve robustness and
  uncertainty,'' \emph{arXiv preprint arXiv:1912.02781}, 2019.

\bibitem{loshchilov2017decoupled}
I.~Loshchilov and F.~Hutter, ``Decoupled weight decay regularization,''
  \emph{arXiv preprint arXiv:1711.05101}, 2017.

\bibitem{torchvision2016}
T.~maintainers and contributors, ``Torchvision: Pytorch's computer vision
  library,'' \url{https://github.com/pytorch/vision}, 2016.

\bibitem{ioffe2015batch}
S.~Ioffe and C.~Szegedy, ``Batch normalization: Accelerating deep network
  training by reducing internal covariate shift,'' in \emph{International
  conference on machine learning}.\hskip 1em plus 0.5em minus 0.4em\relax pmlr,
  2015, pp. 448--456.

\bibitem{ericsson2021well}
L.~Ericsson, H.~Gouk, and T.~M. Hospedales, ``How well do self-supervised
  models transfer?'' in \emph{Proceedings of the IEEE/CVF Conference on
  Computer Vision and Pattern Recognition}, 2021, pp. 5414--5423.

\bibitem{zhao2020makes}
N.~Zhao, Z.~Wu, R.~W. Lau, and S.~Lin, ``What makes instance discrimination
  good for transfer learning?'' \emph{arXiv preprint arXiv:2006.06606}, 2020.

\bibitem{sariyildiz2021concept}
M.~B. Sariyildiz, Y.~Kalantidis, D.~Larlus, and K.~Alahari, ``Concept
  generalization in visual representation learning,'' in \emph{Proceedings of
  the IEEE/CVF International Conference on Computer Vision}, 2021, pp.
  9629--9639.

\bibitem{stuhr2022don}
B.~Stuhr and J.~Brauer, ``Don't miss the mismatch: investigating the objective
  function mismatch for unsupervised representation learning,'' \emph{Neural
  Computing and Applications}, vol.~34, no.~13, pp. 11\,109--11\,121, 2022.

\bibitem{oquab2023dinov2}
M.~Oquab, T.~Darcet, T.~Moutakanni, H.~Vo, M.~Szafraniec, V.~Khalidov,
  P.~Fernandez, D.~Haziza, F.~Massa, A.~El-Nouby \emph{et~al.}, ``Dinov2:
  Learning robust visual features without supervision,'' \emph{arXiv preprint
  arXiv:2304.07193}, 2023.

\bibitem{caron2021emerging}
M.~Caron, H.~Touvron, I.~Misra, H.~J{\'e}gou, J.~Mairal, P.~Bojanowski, and
  A.~Joulin, ``Emerging properties in self-supervised vision transformers,'' in
  \emph{Proceedings of the IEEE/CVF international conference on computer
  vision}, 2021, pp. 9650--9660.

\bibitem{bordes2021high}
F.~Bordes, R.~Balestriero, and P.~Vincent, ``High fidelity visualization of
  what your self-supervised representation knows about,'' \emph{arXiv preprint
  arXiv:2112.09164}, 2021.

\end{thebibliography}

\newpage

% \section{Biography Section}
% If you have an EPS/PDF photo (graphicx package needed), extra braces are
%  needed around the contents of the optional argument to biography to prevent
%  the LaTeX parser from getting confused when it sees the complicated
%  $\backslash${\tt{includegraphics}} command within an optional argument. (You can create
%  your own custom macro containing the $\backslash${\tt{includegraphics}} command to make things
%  simpler here.)
 
% \vspace{11pt}

% \bf{If you include a photo:}\vspace{-33pt}

\begin{IEEEbiography}[{\includegraphics[width=1in,height=1.25in,clip,keepaspectratio]{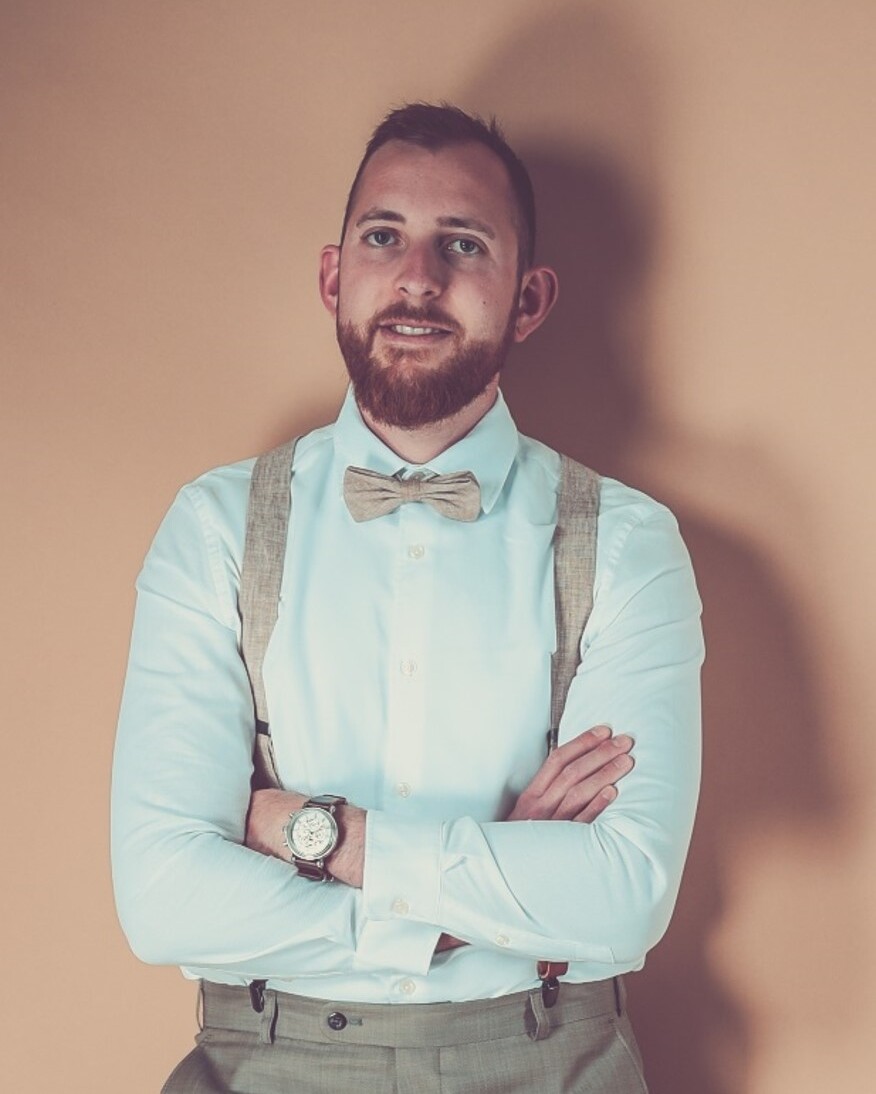}}]{Johann Haselberger}
    received his B.Eng. degree in electrical engineering and information technology and his M.Sc. degree in advanced driver assistance systems from the University of Applied Sciences Kempten, Germany. He is currently working towards his Ph.D. degree in automotive engineering at the Faculty of Mechanical Engineering and Transport Systems, Technical University of Berlin, Germany. Since 2017 he is working as a research assistant at the Institute for Driver Assistance Systems and Connected Mobility at the University of Applied Sciences Kempten, Germany. His main research interests include subject studies on human driving behavior, machine-learning-based driving style modeling, and near-series application of situation-adaptive driving functions.
\end{IEEEbiography}

\begin{IEEEbiography}[{\includegraphics[width=1in,height=1.25in,clip,keepaspectratio]{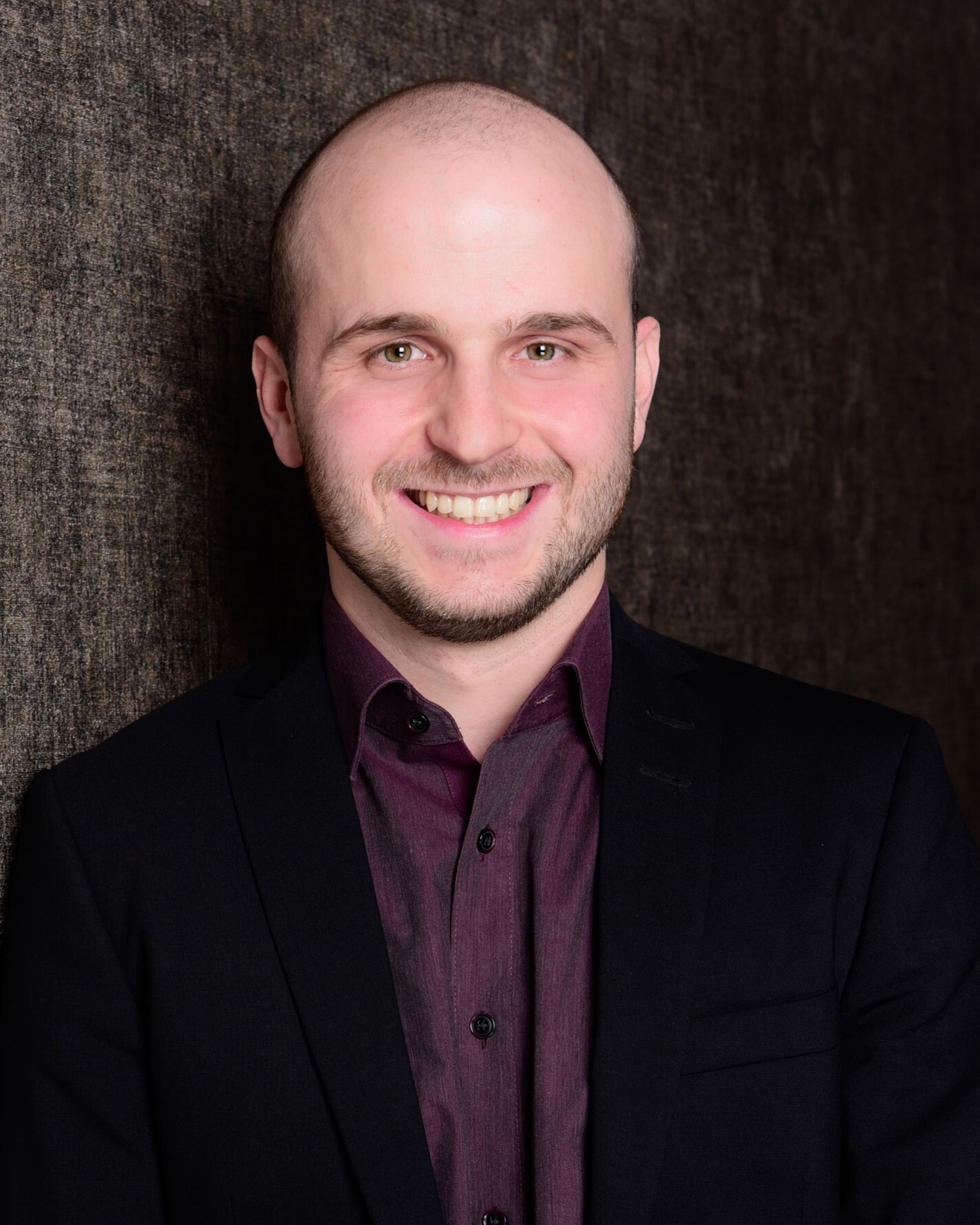}}]{Bonifaz Stuhr}
    received his B.Sc. degree in computer science and his M.Sc. degree in applied computer science from the University of Applied Sciences Kempten, Germany. He holds a Ph.D. in computer science from the Universitat Autònoma de Barcelona, Spain, with an international doctoral research component at the University of Applied Sciences Kempten, Germany. He is currently working as a postdoctoral researcher in artificial intelligence at the Institute for Driver Assistance Systems and Connected Mobility of the University of Applied Sciences Kempten, Germany. His main research interests include neural networks, artificial intelligence, deep learning, unsupervised learning, machine learning, and computer vision.
\end{IEEEbiography}

\begin{IEEEbiography}[{\includegraphics[width=1in,height=1.25in,clip,keepaspectratio]{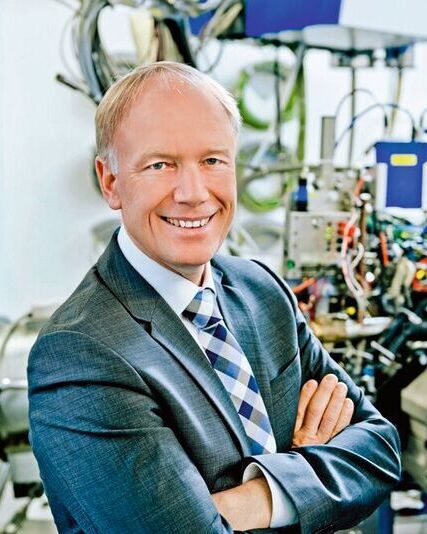}}]{Bernhard Schick}
    received his degree in mechatronic engineering at the University of Applied Sciences Heilbronn. From 1994, whilst at TÜV SÜD, he built up his expertise in the field of vehicle dynamics and advanced driver assistance systems, in various positions up to a general manager. He joined IPG Automotive in 2007 as managing director, where he worked in the field of vehicle dynamics simulation.
    From 2014 he was responsible for calibration and virtual testing technologies as global business unit manager at AVL List, Graz.
    Since 2016, he has been a research professor at the University of Applied Sciences Kempten and the Head of the Institute for Driver Assistance Systems and Connected Mobility. His research focus is automated driving and vehicle dynamics.
\end{IEEEbiography}

\begin{IEEEbiography}[{\includegraphics[width=1in,height=1.25in,clip,keepaspectratio]{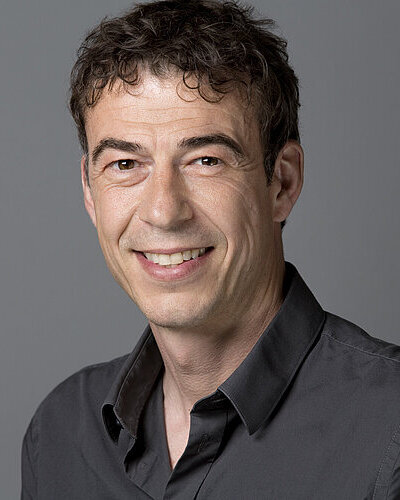}}]{Steffen Müller}
    received the Dipl.-Ing. degree in astronautics and aerospace engineering in 1993 and the Dr.-Ing. degree from Technical University of Berlin in 1998. From 1998 to 2000, he was a project manager at the ABB Corporate Research Center, Heidelberg, Germany. He finished the post-doctoral research at the University of California, Berkeley, in 2001. From 2001 to 2008, he had taken up different leading positions at the BMW Research and Innovation Centre. From 2008 to 2013, he was the founder and leader of the Chair for Mechatronics in Engineering and Vehicle Technology, Technical University of Kaiserslautern, Germany. He is a University Professor and an Einstein Professor with Technical University of Berlin, Germany. He is the Head of the Chair of Automotive Engineering, Faculty of Mechanical Engineering and Transport Systems.
\end{IEEEbiography}

\vfill

\end{document}